\pdfoutput=1

\documentclass[11pt]{article}

\usepackage[preprint]{acl}

\usepackage{times}
\usepackage{latexsym}

\usepackage[T1]{fontenc}

\usepackage[utf8]{inputenc}

\usepackage{microtype}

\usepackage{inconsolata}

\usepackage{graphicx}

\usepackage{amsmath}
\usepackage{tabularx}
\usepackage{booktabs}

\usepackage{colortbl} 
\usepackage{amssymb}
\usepackage{multirow}
\usepackage{fontawesome}
\usepackage{subfigure}
\usepackage{enumitem} 
\usepackage[most]{tcolorbox}

\usepackage{pifont}
\usepackage{xcolor}

\newcommand{\ours}{$\text{LM}_{\text{UniKnow}}$}
\newcommand{\baseline}{KAFT}
\newcommand{\absinst}{Prompting}
\newcommand{\Naive}{Na\"{i}ve}
\newcommand{\naive}{na\"{i}ve}
\newcommand{\retrobust}{RetRobust}
\newcommand{\coiecd}{COIECD}
\newcommand{\coiecda}{$\text{COIECD}_\text{Prompt}$}
\newcommand{\KR}{\texttt{C}}
\newcommand{\UR}{\texttt{E-Only}}
\newcommand{\KI}{\texttt{P-Only}}
\newcommand{\UI}{\texttt{U}}
\newcommand{\framework}{UniKnow}

\title{UniKnow: A Unified Framework for Reliable Language Model Behavior across Parametric and External Knowledge}

\newcommand{\astfootnote}[1]{
    \let\oldthefootnote=\thefootnote
    \setcounter{footnote}{1}
    \renewcommand{\thefootnote}{\fnsymbol{footnote}}
    \footnotetext{#1}
    \let\thefootnote=\oldthefootnote
}

\author{
Youna Kim\textsuperscript{\rm 1},
Hyuhng Joon Kim\textsuperscript{\rm 1},
Minjoon Choi\textsuperscript{\rm 1},
Sungmin Cho\textsuperscript{\rm 1},\\
\textbf{Hyunsoo Cho\textsuperscript{\rm 2},
Sang-goo Lee\textsuperscript{\rm 1 3},
Taeuk Kim\textsuperscript{\rm 4 *}}\\
\textsuperscript{\rm 1}Seoul National University,
\textsuperscript{\rm 2}Ewha Womans University,\\
\textsuperscript{\rm 3}IntelliSys Korea,
\textsuperscript{\rm 4}Hanyang University\\
\texttt{anna9812@europa.snu.ac.kr} \quad \texttt{kimtaeuk@hanyang.ac.kr}\\
}

\begin{document}
\maketitle

    \begin{abstract}

Language models often benefit from external knowledge beyond parametric knowledge.
While this combination enhances performance, achieving reliable knowledge utilization remains challenging, as it requires assessing the state of each knowledge source based on the presence of relevant information.
Yet, prior work on knowledge integration often overlooks this challenge by assuming ideal conditions and provides limited coverage of knowledge scenarios.
To address this gap, we introduce \textbf{\framework}, a \textbf{Uni}fied framework for reliable LM behavior across parametric and external \textbf{Know}ledge. 
\framework\ enables controlled evaluation across knowledge scenarios such as knowledge conflict, distraction, and absence conditions that are rarely addressed together.
Beyond evaluating existing methods under this setting, we extend our work by introducing \framework-Aware methods to support comprehensive evaluation.
Experiments on \framework\ reveal that existing methods struggle to generalize across a broader range of knowledge configurations and exhibit scenario-specific biases. 
\framework\ thus provides a foundation for systematically exploring and improving reliability under knowledge scenarios.

\astfootnote{Corresponding author.}
\end{abstract}

\section{Introduction}

Language models (LMs), trained on large-scale corpora, exhibit the capacity to address a broad range of tasks by leveraging their pre-trained parametric knowledge \citep{grattafiori2024llama3herdmodels, qwen2.5}.
However, LMs are confined to the static pre-trained knowledge and therefore struggle to handle tasks requiring information beyond this boundary, such as long-tail \citep{pmlr-v202-kandpal23a, mallen-etal-2023-trust} or time-sensitive information \citep{pmlr-v162-liska22a}.
To overcome these limitations, LMs often benefit from dynamically incorporating external knowledge, commonly through retrieval-augmented generation (RAG), thereby granting access to up-to-date, task-relevant information at inference time \citep{chen-etal-2017-reading, asai-etal-2023-retrieval}.

\begin{figure}[t]
\begin{center}
    \includegraphics[width=1\columnwidth]{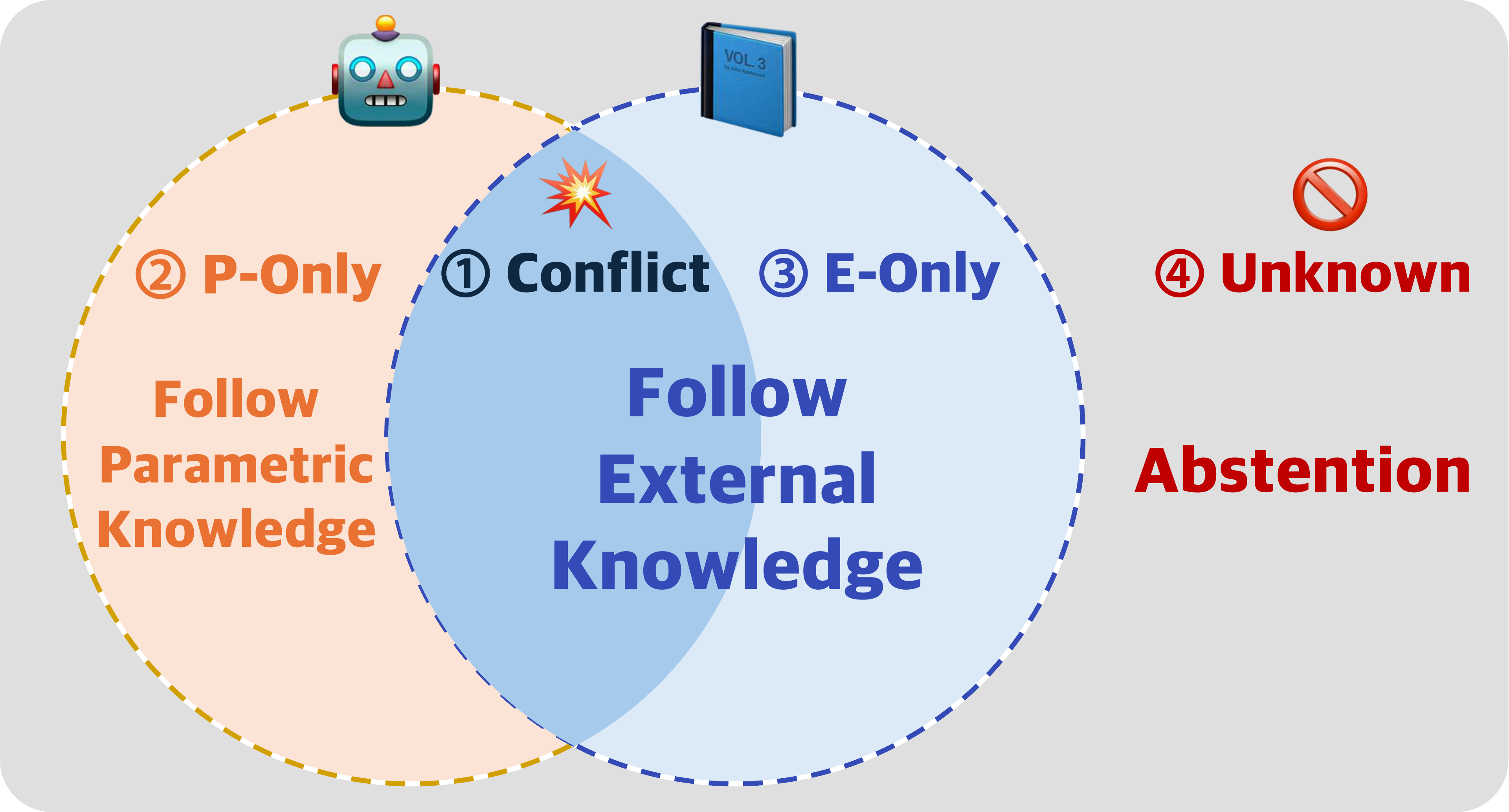}
      \caption{Four knowledge scenarios in \framework\ are defined by the boundaries of parametric and external knowledge sources. Each region illustrates the expected LM behavior for each scenario.}
      \label{fig:main_figure3}
\end{center}
\end{figure}

The integration of parametric and contextual knowledge has broadened the capabilities of LMs, driving their application in knowledge-intensive and sensitive domains \citep{tsatsaronis2015overview, jin2019pubmedqa, dasigi-etal-2021-dataset}.
Consequently, the reliability of LMs has become a vital consideration \citep{wen-etal-2024-characterizing}, with models expected to not only recognize the boundaries of their possessed knowledge but also identify when relevant information is missing.
While prior work has tackled various dimensions of knowledge integration \citep{su2024textttconflictbank, yoran2023making}, these studies have typically remained fragmented, 
providing an incomplete assessment of reliability \citep{li-etal-2023-large, cheng2024understandinginterplayparametriccontextual}.
Moreover, knowledge utilization methods developed under such narrow environments still lack validation in more realistic and compositional knowledge scenarios.

To this end, we introduce \textbf{\framework}, a unified framework for reliable LM behavior across parametric and external knowledge.
While reliability may encompass a broader range of factors, this work focuses on the presence of \textit{relevant} information in each parametric and external knowledge source.
Central to \framework\ is the notion of \textit{relevance}, which we define as whether a knowledge source provides sufficient and contextually supporting information to answer a query.
For example, when asked “Who is the president of the United States?”, an LM might answer "Biden" based on its parametric knowledge, while the context might refer to "Trump"--both are considered relevant.

\framework\ is designed to categorize and assess four distinct scenarios as illustrated in Figure \ref{fig:main_figure3}: (1) Conflict, (2) Parametric-Only, (3) External-Only, and (4) Unknown.
When only a single relevant source is available, the model is expected to ground its output solely in that source.
Furthermore, if both sources are relevant but conflicting (1), the model should prioritize the external knowledge, as it generally offers more up-to-date and task-specific information.
If neither source provides relevant knowledge (4), the model should recognize its limitations and abstain from generating hallucinations \citep{zhang-etal-2024-r, feng-etal-2024-dont}.

To examine how existing methods developed under partial scenario coverage generalize to \framework, we evaluate two \naive\ baselines and three existing methods with different scenario coverage. 
We further complement this evaluation by introducing \framework-Aware methods--inference-based and training-based--covering all scenarios in \framework\ by explicitly incorporating relevance-based knowledge conditions into their formulation.

Our in-depth analysis under \framework\ reveals that methods appearing reliable in individual scenarios often fail in composite scenarios requiring simultaneous consideration of both knowledge sources.
We further uncover how LM behavior shifts across scenarios, highlighting biases specific to scenario types.
Notably, training with \framework-aligned supervision significantly improves reliability. 
Together, these findings enable a more comprehensive understanding of LM alignment potential under \framework\ and mark a substantial step toward bridging the gap between narrow knowledge settings and a unified framework.

    \section{Related Works}

\paragraph{Knowledge Conflict}
Parametric knowledge is inherently static, whereas external knowledge can be delivered in response to diverse circumstances.
This dynamic provision often results in discrepancies between the parametric memory and the external context. 
Studies have examined the conflict through the lens of external knowledge features, such as temporal shifts \citep{NEURIPS2023_9941624e, dhingra-etal-2022-time}, synthetically updated facts \citep{longpre-etal-2021-entity}, and contextual plausibility \citep{xie2023adaptive, tan-etal-2024-blinded}.
Building on these findings, several approaches aim to improve external knowledge incorporation, primarily through contrastive decoding \citep{shi-etal-2024-trusting, jin-etal-2024-cutting, yuan-etal-2024-discerning}.
Yet many existing approaches \citep{liu-etal-2024-untangle, wang2024resolving, jin-etal-2024-tug} still treat any mismatch between model output and context as a conflict, often neglecting whether the model had prior access to that information.

\paragraph{Robustness against Irrelevance}
Although external knowledge is intended to supply LM's knowledge, in real-world scenarios (i.e. RAG), it may not always be relevant. 
LMs face challenges in handling irrelevant context, which often leads to performance degradation \citep{shen-etal-2024-assessing}. 
RAG is particularly susceptible, as retrieval errors can introduce a misleading but plausible context \citep{wu2024how}.
To mitigate this, researchers have explored methods to encourage LMs to rely on parametric knowledge when external information is irrelevant--either at inference time \citep{yu-etal-2024-chain, park2024enhancing, baek-etal-2023-knowledge-augmented-language} or through training \citep{yoran2023making, asai2024selfrag, xia2024improving, luo-etal-2023-search}.
Despite their effectiveness at mitigating the influence of irrelevant context, these approaches entirely overlook the presence of relevant information in their parametric knowledge.

\begin{figure*}
\centering
\includegraphics[width=1\linewidth]{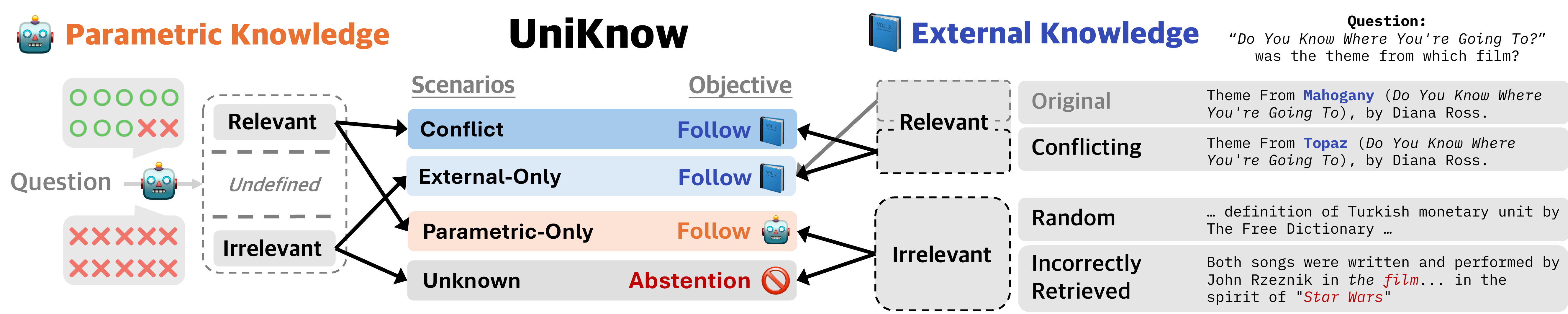}
\caption{Overview of \framework.}
\label{fig:framework_figure}
\end{figure*}

\paragraph{Abstention}
A growing line of work focuses on aligning LMs to abstain when appropriate--specifically when the model does not possess the relevant knowledge--to prevent hallucination and ensure reliable model behavior \citep{wen2025knowlimitssurveyabstention}.
Some approaches quantify uncertainty \citep{10820047, kuhn2023semantic, kadavath2022languagemodelsmostlyknow} in parametric knowledge and relabel training data accordingly to guide abstention behavior \citep{feng-etal-2024-dont, zhang-etal-2024-r, wen2024know}.
Recently, studies have begun to explore abstention based on the relevance of external knowledge \citep{wen-etal-2024-characterizing, kim2025speakabstaincontrastivedecoding}.

\paragraph{Knowledge Frameworks}
There have been efforts to unify various aspects of knowledge utilization to understand LM behaviors.
\citet{li-etal-2023-large} trains LMs to generate either parametric- or context-grounded responses depending on the context type, whereas \citet{neeman-etal-2023-disentqa} trains LMs to generate both in parallel.
Similar to our work, \citet{cheng2024understandinginterplayparametriccontextual} proposes a benchmark to investigate whether LMs can express possessed parametric knowledge when exposed to various context types.
While prior approaches have provided valuable insights into how LMs utilize knowledge, we extend this perspective with a framework centered on reliability.
We offer a more comprehensive view by jointly considering the relevance of information within both parametric and external knowledge, and by addressing conflict, irrelevance, and abstention within a unified framework.

\section{\framework}
This work focuses on context-augmented generation in open-domain question-answering, facilitating LMs to leverage their \textbf{parametric} knowledge while simultaneously utilizing \textbf{external} knowledge to answer a given query $q$.
This section first defines each knowledge source based on the availability of relevant information.
Guided by this taxonomy, we introduce \textbf{\framework}, a \textbf{Uni}fied framework for reliable LM behavior across parametric and external \textbf{Know}ledge, covering four distinct scenarios as illustrated in Figure \ref{fig:framework_figure}.
We then describe the construction process of estimating the parametric knowledge and designing diverse context types.

\subsection{Definition of Knowledge Sources}

\textbf{Parametric knowledge (PK)} refers to information encoded in an LM during pretraining. 
Since this knowledge is bound by its pretraining data, 
we define that relevant information resides in PK ($\exists_{\text{PK}}$) if $\text{LM}(\hat{a} \mid q) = a^{*}_{\text{PK}}$, where $a^{*}_{\text{PK}}$ denotes the answer grounded in the LM's pretraining data \citep{bang2025hallulensllmhallucinationbenchmark}.
Still, PK remains inherently static and may not align with the most recent world knowledge.

\textbf{External knowledge (EK)} indicates any information provided at inference time as the input context. 
To solely evaluate the LM's ability to utilize relevant knowledge, we assume that all provided EK is factually aligned with world knowledge.
Under this assumption, we assess LM behavior in both relevant ($\exists_{\text{EK}}$) and irrelevant ($\varnothing_{\text{EK}}$) contexts.

\subsection{Scenarios in \framework}
\label{sec:knowledge-handling_scenarios}

\framework\ is designed to cover all possible scenarios regarding the presence of relevant PK and EK.
This gives rise to four distinct scenarios, each reflecting real-world challenges such as conflict resolution, over-reliance, and hallucination risk.
Since each challenge has its own expected behavior, we define scenario-specific expectations as follows.

\begin{itemize}[topsep=5pt,leftmargin=10pt]
    \item \textbf{Conflict (\KR)}: 
    $(\exists_{\text{PK}}, \exists_{\text{EK}}) \text{ and } a^{*}_{\text{PK}} \ne a^{*}_{\text{EK}}$ \\
    The conflict between knowledge sources arises when EK presents relevant information contradicting what LM knows \citep{xu-etal-2024-knowledge-conflicts}.
    While PK and EK may either align or conflict, we focus on the latter, allowing us to evaluate whether LMs can correctly prioritize EK. 
    \item \textbf{External-Only (\UR)}:
    $(\varnothing_{\text{PK}}, \exists_{\text{EK}})$ \\
    The model lacks PK with relevant information and is expected to rely on relevant EK.
    \item \textbf{Parametric-Only (\KI)}: 
    $(\exists_{\text{PK}}, \varnothing_{\text{EK}})$ \\
    The model is required to rely on its PK with relevant information and ignore irrelevant EK.
    \item \textbf{Unknown (\UI)}: 
    $(\varnothing_{\text{PK}}, \varnothing_{\text{EK}})$ \\
    Neither knowledge source is sufficient, and the model is expected to abstain from answering.
\end{itemize}

\subsection{Parametric Knowledge Estimation}
\label{dataset:parametric-knowledge}

We estimate the presence of relevant PK by assessing whether the LM is capable of generating a correct answer to a given $q$ without access to external context.
Following prior works, we assess the factual \textit{correctness} \citep{zhang-etal-2024-r, zhang-etal-2024-self, wang-etal-2024-factuality} and \textit{consistency} \citep{kuhn2023semantic, 10820047, amayuelas-etal-2024-knowledge} of the prediction utilizing its PK.
We classify $q$ as $\exists_{\text{EK}}$ if both conditions are satisfied, and as $\varnothing_{\text{PK}}$ otherwise.

For each $q$, we sample $n$ responses using $q$ alone: $a_{i} \sim LM(a \mid q)$ for $i = 1, .., n$.
If the proportion of correct responses is greater than or equal to the threshold $\tau$, we classify $q$ as $\exists_{\text{EK}}$:
\begin{equation}
\frac{1}{n} \sum_{i=1}^{n} \mathbf{1}[a_{i} = a^{*}_{\text{PK}}] \geq \tau \Rightarrow q \in \exists_{\text{PK}}
\end{equation}
If none of the responses are correct, we assign $q \in \varnothing_{\text{PK}}$.
Questions falling between these thresholds are considered \textit{undefined} and excluded from scenario construction.
We set $n=10$ and $\tau = 0.7$ in our implementation.

\subsection{External Knowledge Construction}
\label{section:externalKnowledge}

To operationalize each scenario, we construct context types tailored to diverse conditions.
In addition to the original context, we construct conflicting and two types of irrelevant contexts: 
(1) topically unrelated random contexts, and (2) incorrectly retrieved contexts with high retriever score.
This allows fine-grained control over the degree of relevance, capturing challenges ranging from knowledge conflicts to misleading but plausible distractors.

\paragraph{Relevant contexts}
The \textit{original} context refers to the context paired with the question-answer pair in the dataset. 
We derive a \textit{conflicting} context by providing {\scshape Llama 3 70B Instruct} \citep{grattafiori2024llama3herdmodels} with the original context and the corresponding answer to generate an alternative answer while preserving its part of speech.
The original answer span is then replaced with the conflicting answer, introducing an intended conflict with the model's PK.
Note that in \KR\ scenario, we use only conflicting contexts, whereas \UR\ scenario includes both original and conflicting contexts.

\paragraph{Irrelevant contexts}
We consider two key aspects for irrelevant context selection: the absence of the answer span (i.e., uninformative) and the potential semantic relevance \citep{wu2024how} that may mislead the model (i.e., misleading).
To capture both uninformative and misleading cases, we include two types of contexts.
A \textit{randomly} sampled context from the same dataset, topically unrelated to the question, and not containing the original answer.
The \textit{incorrectly retrieved contexts} also lack the answer but may appear topically relevant, thereby creating a false sense of relevance.
We obtain these incorrectly retrieved contexts by querying a Wikipedia corpus using the {\scshape Contriever-msmarco} retriever \citep{izacard2022unsupervised}, and then select the highest-ranked context that does not contain the answer.
This setting captures challenges in real-world RAG, where retrieval often returns plausible but irrelevant information.

    \definecolor{g}{HTML}{27b376}
\definecolor{r}{HTML}{d73027}

\begin{table}[t]
\centering
\setlength{\tabcolsep}{12pt}
\resizebox{\columnwidth}{!}{
\begin{tabular}{lcccc}
\toprule

\textbf{Methods} & \textbf{Conflict} & \textbf{E-Only} & \textbf{P-Only} & \textbf{Unknown} \\
\midrule
\coiecd        & {\color{g} \ding{52}} & {\color{g} \ding{52}} & {\color{r} \ding{56}} & {\color{r} \ding{56}} \\
\retrobust    & {\color{r} \ding{56}} & {\color{g} \ding{52}} & {\color{g} \ding{52}} & {\color{r} \ding{56}} \\
\baseline         & {\color{g} \ding{52}} & {\color{g} \ding{52}} & {\color{g} \ding{52}} & {\color{r} \ding{56}} \\
\coiecda        & {\color{g} \ding{52}} & {\color{g} \ding{52}} & {\color{g} \ding{52}} & {\color{g} \ding{52}} \\
\ours   & {\color{g} \ding{52}} & {\color{g} \ding{52}} & {\color{g} \ding{52}} & {\color{g} \ding{52}} \\
\bottomrule
\end{tabular}}
\caption{Comparison of methods based on their consideration of each \framework\ scenarios. `{\color{g} \ding{52}}' indicates the method explicitly accounts for the corresponding scenario, while `{\color{r} \ding{56}}' denotes that it does not.}
\label{tab:method_comparison}
\end{table}

\section{Knowledge Utilization Methods}

This section describes the methods used to evaluate model behavior under \framework.
As an initial baseline, we take a \textbf{prompting} approach, instructing the model to consider the presence of knowledge sources for a reliable generation.
We also perform \textbf{na\"{i}ve} greedy generation with a simple QA template.
We further assess three existing methods alongside two \framework-aware approaches.
Table \ref{tab:method_comparison} summarizes the scenario coverage of each method.
Implementation details of the approaches are presented in Appendix~\ref{app:methods}.

\subsection{Existing Methods}
We adapt three existing methods, each designed to handle only partial scenarios of \framework.

\paragraph{Conflict}
Methods for resolving knowledge conflict aim to overwrite the model's PK with EK.
To this end, context-aware contrastive decoding approaches have been widely explored \citep{shi-etal-2024-trusting, zhao-etal-2024-enhancing}.
Among them, we utilize \textbf{\coiecd}\footnote{Contextual Information-Entropy Constraint Decoding} \citep{yuan-etal-2024-discerning}, a state-of-the-art method that amplifies the context-informed distribution when conflict arises.

\paragraph{Parametric-Only}
\textbf{\retrobust} \citep{yoran2023making} fine-tunes the LM with augmented training data, incorporating irrelevant context alongside the original context. 
The goal of \retrobust\ is to improve its robustness against irrelevant contexts.

\paragraph{Conflict and Parametric-Only}
\citealp{li-etal-2023-large} also adopts a fine-tuning approach, \textbf{\baseline}, a knowledge-aware fine-tuning that addresses both knowledge conflict and irrelevance.
Their training data includes original, conflicting, and irrelevant contexts, aiming to improve the LM's overall ability to utilize external knowledge effectively.

\subsection{\framework-Aware Methods}

To evaluate the impact of covering all \framework\ scenarios on model behavior, we construct two \framework-Aware methods.

\paragraph{\framework-Aware Inference}
To explicitly account for \framework's scenarios during inference, we introduce \textbf{\coiecda}, an extension of \coiecd\ that additionally incorporates prompting into the decoding process.
By explicitly considering all the possible scenarios, we expect \coiecda\ to cover a broader range of cases.

\paragraph{\framework-Aware Training}
We investigate whether reliability can be improved by training LMs with supervision aligned to knowledge scenarios defined in \framework.
We design scenario-aware training data that explicitly reflects the presence or absence of relevant information in both knowledge sources.
The key lies in the scenario-aware construction of the training data.

To prepare training data, we sample a balanced set of $q \in \exists_{\text{PK}}$ and $q \in \varnothing_{\text{PK}}$, as determined by the criteria in Section \ref{dataset:parametric-knowledge}.
As illustrated in Figure~\ref{fig:train_data}, each $q$ is then paired with four types of external contexts described in Section \ref{section:externalKnowledge} to cover knowledge scenarios.
To maintain the LM's ability to answer when the context contains information that matches with its PK ($a^{*}_{\text{PK}} = a^{*}_{\text{EK}}$), we include the original context paired with $q \in \exists_{\text{PK}}$ during training, although it is excluded from the \KR\ scenario analysis.
For scenarios where relevant information is available--\KR, \UR, and \KI--\ours\ is optimized to produce the expected answer corresponding to each scenario.
In the \UI\ scenario, \ours\ is trained to abstain by generating "\texttt{unknown}".

\section{Experimental Setting}

\subsection{Implementation Details}

\paragraph{Datasets}
\framework\ employs seven QA datasets from diverse knowledge domains: NaturalQuestions (NQ), TriviaQA, HotpotQA, SQuAD, BioASQ, TextbookQA, and RelationExtraction (RE) \citep{kwiatkowski-etal-2019-natural, joshi2017triviaqalargescaledistantly, yang2018hotpotqadatasetdiverseexplainable, rajpurkar2016squad100000questionsmachine, tsatsaronis2015overview, Kembhavi_2017_CVPR, levy2017zeroshotrelationextractionreading}. 
We use the dataset versions curated by the Machine Reading for Question Answering (MRQA) benchmark \cite{fisch2019mrqa2019sharedtask}. 
As the impact of context length is beyond the scope of our study, we limit the context to approximately 100 words to ensure experimental control. 

\paragraph{Models}
We use open-source auto-regressive language models, including {\scshape Llama2} (7B \& 13B, \citealp{touvron2023llama}), {\scshape Llama3-8B} \citep{grattafiori2024llama3herdmodels}, {\scshape Mistral-7B v0.3} \citep{jiang2023mistral}, and {\scshape Qwen 2.5} (1.5B \& 3B \& 7B \& 14B, \citealp{qwen2.5}).
Training-based methods are evaluated in a zero-shot setting, whereas inference-only methods utilize two-shot demonstrations.
More details on datasets and templates are in Appendix \ref{appendix:implementationDetails}.

\begin{figure}[t]
\begin{center}
    \includegraphics[width=1\columnwidth]{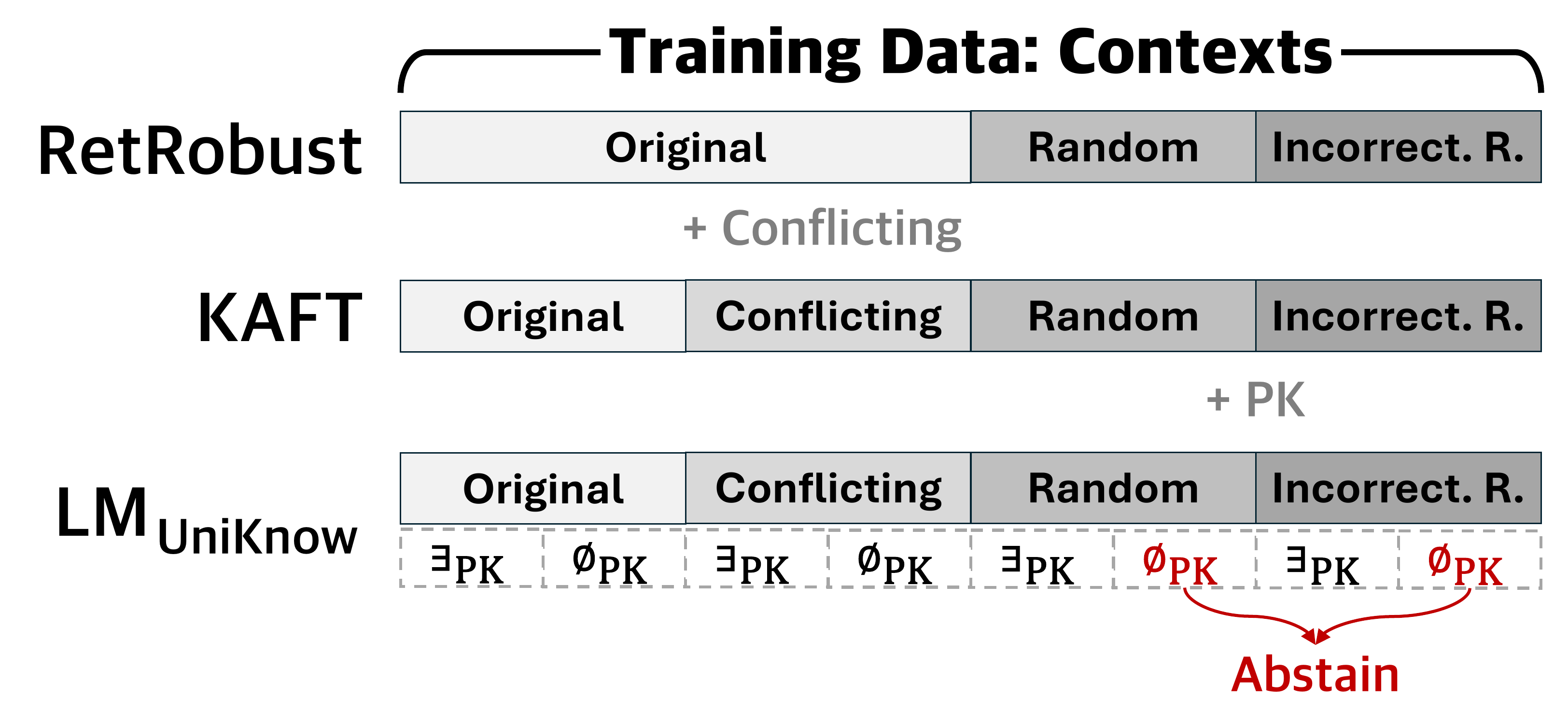}
      \caption{ 
      Comparison of training-based methods on their training data.
      }
      \label{fig:train_data}
\end{center}
\end{figure}

\paragraph{Training Details}
For a fair comparison, all training-based methods share the same settings.
Utilizing the training set of NQ and TriviaQA, we randomly sample 250 questions from each of $\exists_{\text{PK}}$ and $\varnothing_{\text{PK}}$, resulting in a total of 1,000 samples.
As illustrated in Figure \ref{fig:train_data}, we pair each $q$ with four context types, resulting in 4,000 question-context pairs.
In case of \retrobust, since it does not use conflicting contexts, we additionally sample 1,000 questions and pair them with the original context.
QLoRA \citep{dettmers2023qlora} is applied for efficient training.
Appendix~\ref{app:hp_train} provides additional training details.

\begin{figure*}
\centering
\includegraphics[width=\linewidth]{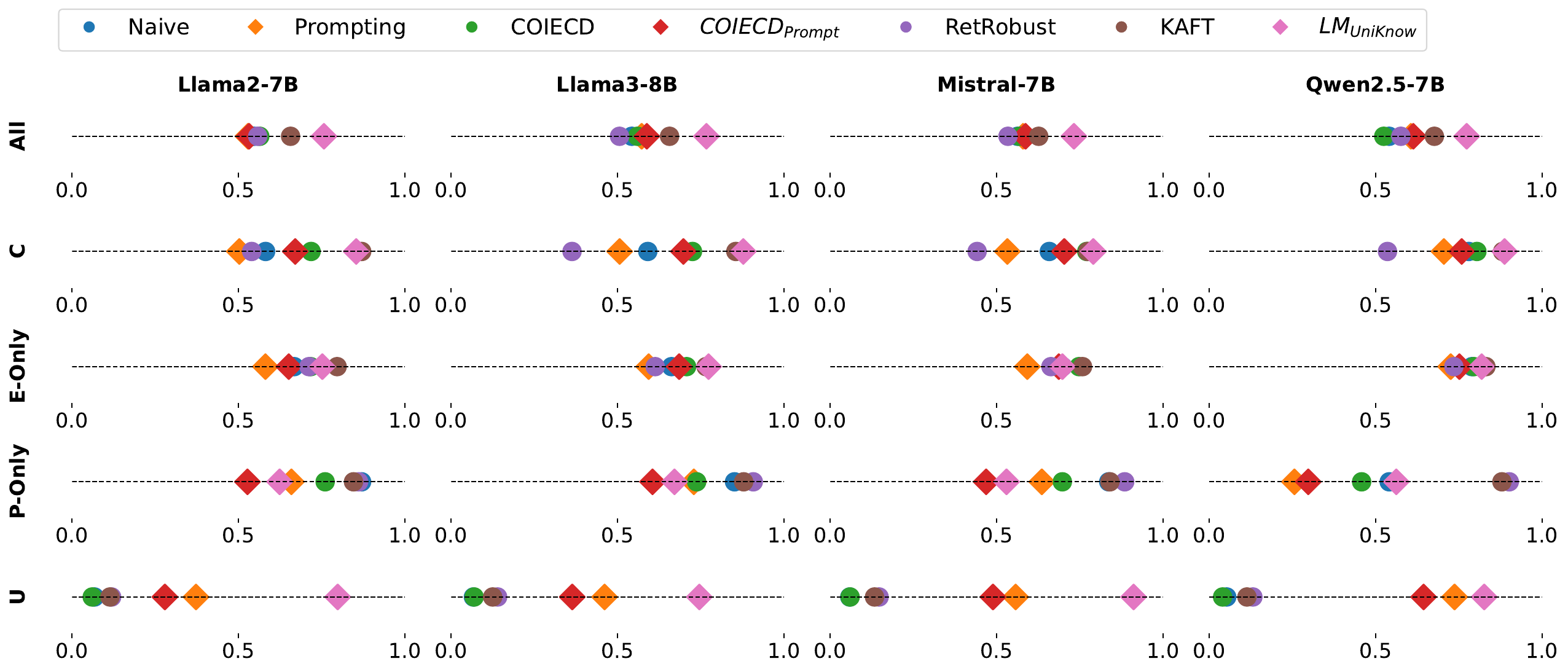}
\caption{EM scores by scenario and model. \texttt{All} indicates scores averaged across all scenarios. Methods marked with diamonds incorporate abstention, while those with circle markers do not.}
\label{fig:scenario}
\end{figure*}

\subsection{Evaluation Metrics}

We use Exact Match (EM) to assess whether the model's prediction aligns with the expected answer.
Note that the expected answer for each scenario differs, as defined in Section \ref{sec:knowledge-handling_scenarios}.
Still, it is equally important to evaluate LM behavior on samples that are \textit{undefined} with respect to the presence of relevant PK.
To reflect more realistic usage settings, we also evaluate the full samples within \framework\ and report the accuracy (\texttt{Acc}) and reliability (\texttt{Rely}) score \citep{xu2024rejection}, which captures both correctness and appropriate abstention.
These are computed based on the number of correct ($N_c$), incorrect ($N_i$), and abstained ($N_a$) responses with EM.\footnote{$N$: The total number of responses.}

\texttt{Rely} score captures the balance of two components: \texttt{Acc} ($\frac{N_c}{N}$) and truthfulness (\texttt{Truth}).
\texttt{Truth} measures the proportion of responses that are either correct or abstained ($\frac{N_c + N_a}{N}$), thereby ensuring that the model avoids generating incorrect outputs.
To discourage excessive abstention, the answer rate ($\texttt{Ans} = \frac{N_c + N_i}{N}$) is used as a weighting factor. 
\texttt{Rely} is computed as $\texttt{Ans} \times \texttt{Truth} + (1 - \text{Ans}) \times \texttt{Acc}$.
\texttt{Rely} is high when LM provides correct answers and abstains appropriately, while penalizing both incorrect outputs and excessive abstention.

\begin{figure}
\centering
\includegraphics[width=\columnwidth]{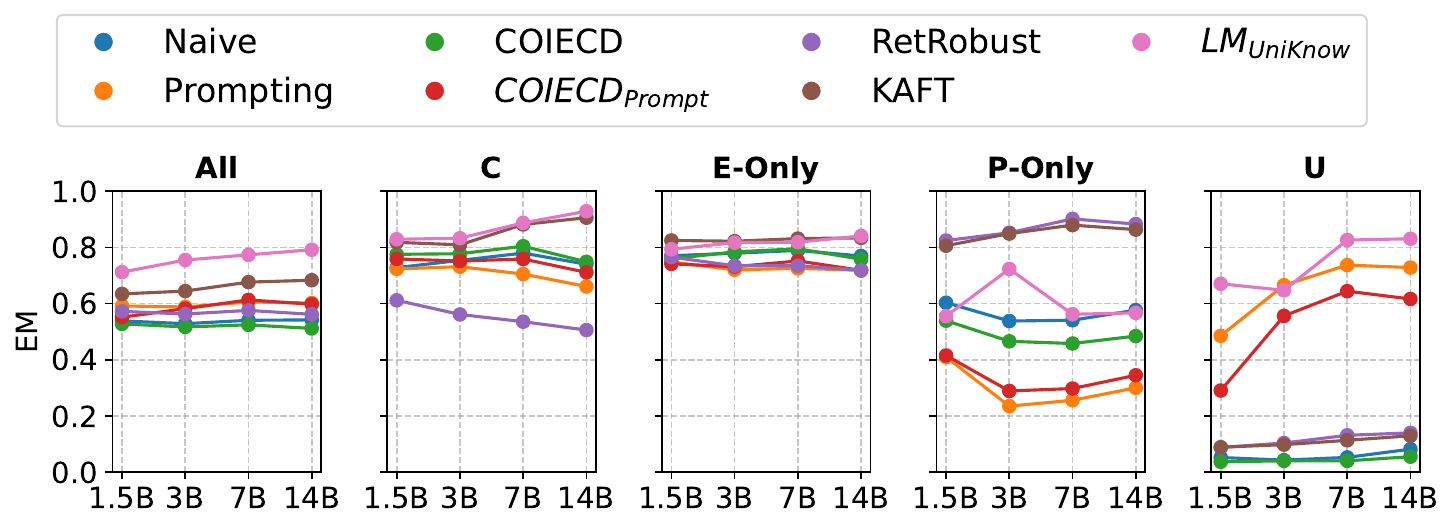}
\caption{EM scores of {\scshape Qwen} models across different sizes, shown by scenario.}
\label{fig:qwen_size}
\end{figure}

\section{Results on \framework}

\subsection{Main Results}
\label{sec:main_results}

Figure \ref{fig:scenario} illustrates the performance across the four \framework\ scenarios, \KR, \UR, \KI, and \UI, and the overall averaged performance (\texttt{All}).
To assess generalization across knowledge domains, we report EM scores averaged over all datasets, comprising two in-domain and five out-of-domain sets for training-based methods.

\paragraph{Broader scenario coverage leads to better overall results.}
\ours, which covers all scenarios, achieves the best overall performance, followed by \baseline.
Other methods, designed with a subset of scenarios, lead to limited performance gains, often falling below or only marginally above \naive.
Meanwhile, \coiecda\ consistently outperforms both \coiecd\ and \absinst\ in three out of four models, demonstrating the extensibility potential of existing methods. 

\paragraph{Resolving conflicts with known knowledge poses a greater challenge than simply incorporating new, unknown information.}
Compared to \KR\ scenario, the performance points in \UR\ are more tightly clustered with less variance. 
It demonstrates that LM behavior is influenced not only by context type itself, but also by its interaction with PK.
Still, a similar trend is observed across methods in both \KR\ and \UR\ scenarios.
Notably, the performance drop of \retrobust\ is more pronounced in the \KR\ scenario than in \UR, reflecting its limited ability to handle contradictory information effectively.

\paragraph{A trade-off between answering and abstention arises under irrelevant contexts.}
Methods that prioritize answerability without accounting for the presence of PK, such as \coiecd, \retrobust, and \baseline, achieve strong performance in \KI\ scenario.
However, in \UI\ scenario, they are more likely to generate hallucinations.
In contrast, methods that incorporate abstention ability, including \absinst, \coiecda, and \ours, handle \UI\ with abstention behavior, but suffer in a trade-off of exhibiting lower performance in \KI.
Among these, \ours\ demonstrates the largest performance gain in \UI\ scenario, driven by its consideration of the model's knowledge state.

\paragraph{Larger LMs generally improve reliability, with distinct trends across scenarios.}
Based on Figure \ref{fig:qwen_size}, the performance in \UR\ scenario remains relatively unaffected by scale, suggesting that EK utilization does not strongly benefit from larger LMs.
In \KR\ and \KI\ scenarios, gains depend on whether the method is explicitly trained for those conditions.
By contrast, in \UI\ scenario, abstention performance improves consistently with scale, indicating that larger LMs are better at recognizing knowledge limitations and abstaining accordingly.

\subsection{Error Analysis}

\begin{figure}[t]
    \centering
    \includegraphics[width=1\columnwidth]{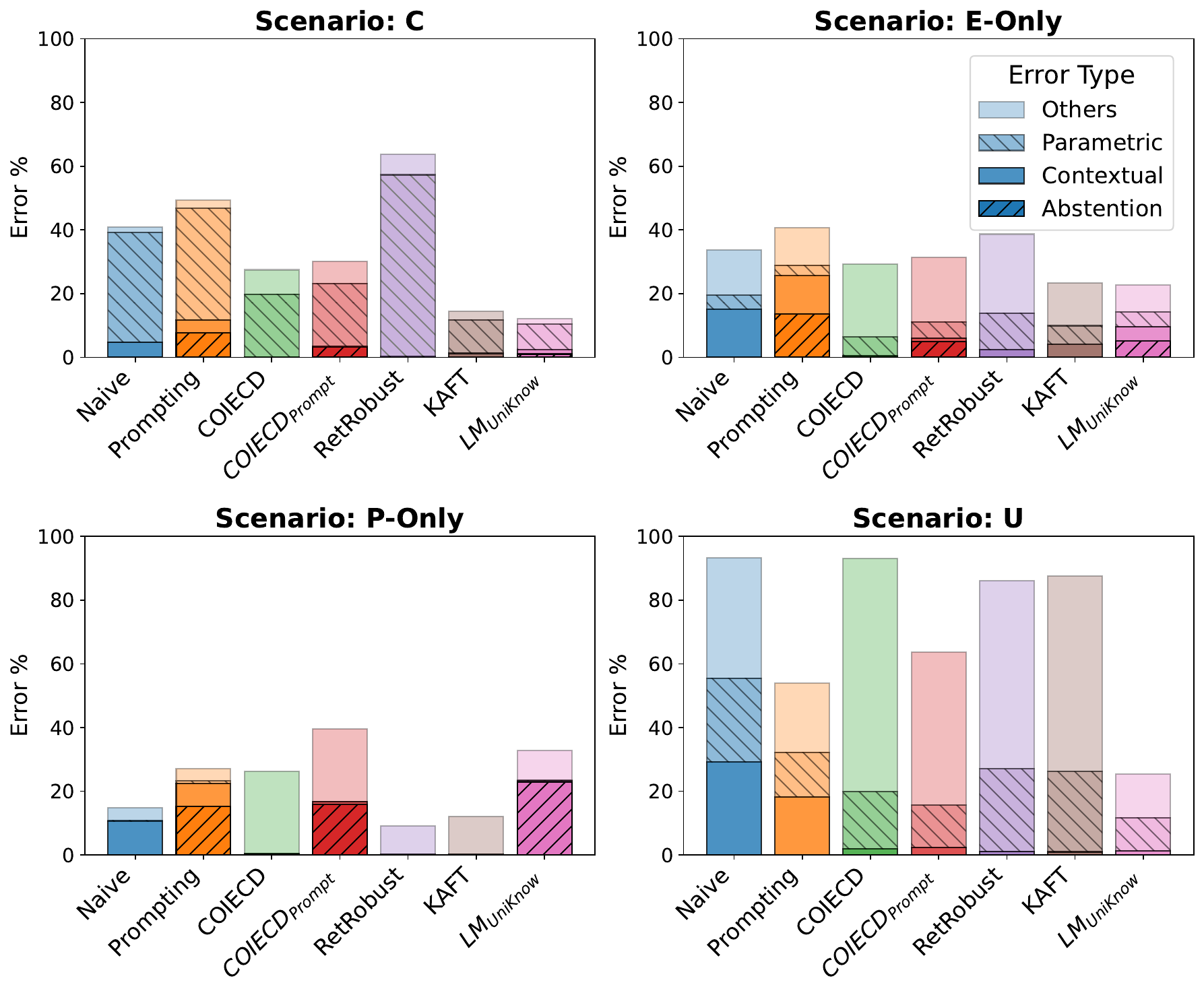} 
    \caption{Stacked error type distributions across methods for each knowledge scenario. Transparency reflects error type. Evaluated using {\scshape Llama3-8B}.}
    \label{fig:error_analysis}
\end{figure}

Since LMs may exhibit scenario-specific biases, we analyze output errors to examine such patterns in detail.
Incorrect responses are categorized into four types: contextual, parametric, false abstention, and others.
\textit{Contextual errors} occur when the model generates an incorrect response grounded on the given context.
In case of relevant context, this involves extracting incorrect information; in the case of irrelevant content, the model is misled by unrelated content.
\textit{Parametric errors} refer to errors generated based on the model's PK.
In the \KR\ scenario, this reflects the model's failure to follow the given context, exhibiting a parametric bias. 
\textit{False abstention} is counted as an error in three scenarios where the model possesses at least one relevant knowledge, except \UI.
\textit{Other} includes incorrect responses that do not fall into the above categories. 
Figure \ref{fig:error_analysis} shows the error distribution for Llama 3 8B across the four knowledge-handling scenarios.

\paragraph{Over-reliance on PK depends on the presence of PK.}
In the \KR\ scenario, where the model possesses the relevant information, all methods exhibit the highest rate of parametric errors compared to other error types.
In contrast, such error is much less common in \UR\ scenario.
Even with \coiecd, which explicitly targets knowledge conflict, the rate of parametric error remains significantly higher in \KR\ than in \UR.
Unlike prior works that focus solely on controlling EK via conflicting contexts, our findings highlight that over-reliance becomes more evident when scenarios are further distinguished by the presence of PK.

\paragraph{Contextual errors are rare across most methods, except for \naive\ approaches.}
In \naive\ approaches, contextual errors are observed in all scenarios, particularly in \UR\ and \UI.
This indicates that when the required knowledge is absent from the model's parametric memory, it tends to rely on the provided context but often fails to utilize it correctly (\UR) or is misled by irrelevant information (\UI).
In contrast, most other methods effectively mitigate context misinterpretation, as evidenced by the near absence of contextual errors.

\paragraph{Abstention error occurs most frequently in \KI\ scenario, while it is rare under relevant contexts.}
Methods guided to abstain appropriately tend to exhibit relatively high abstention bias in \KI.
This again highlights the importance of the trade-off mitigation.
Interestingly, the abstention error rate of \coiecda\ remains comparable to that of \baseline\ in \KI, but is significantly reduced in \UR.
This indicates that combining the strengths of \coiecd\ and \absinst\ leads to more proper abstention behavior across scenarios.

\section{Additional Analysis on Reliability} \label{sec:additional}

\begin{figure}
\centering
\includegraphics[width=\columnwidth]{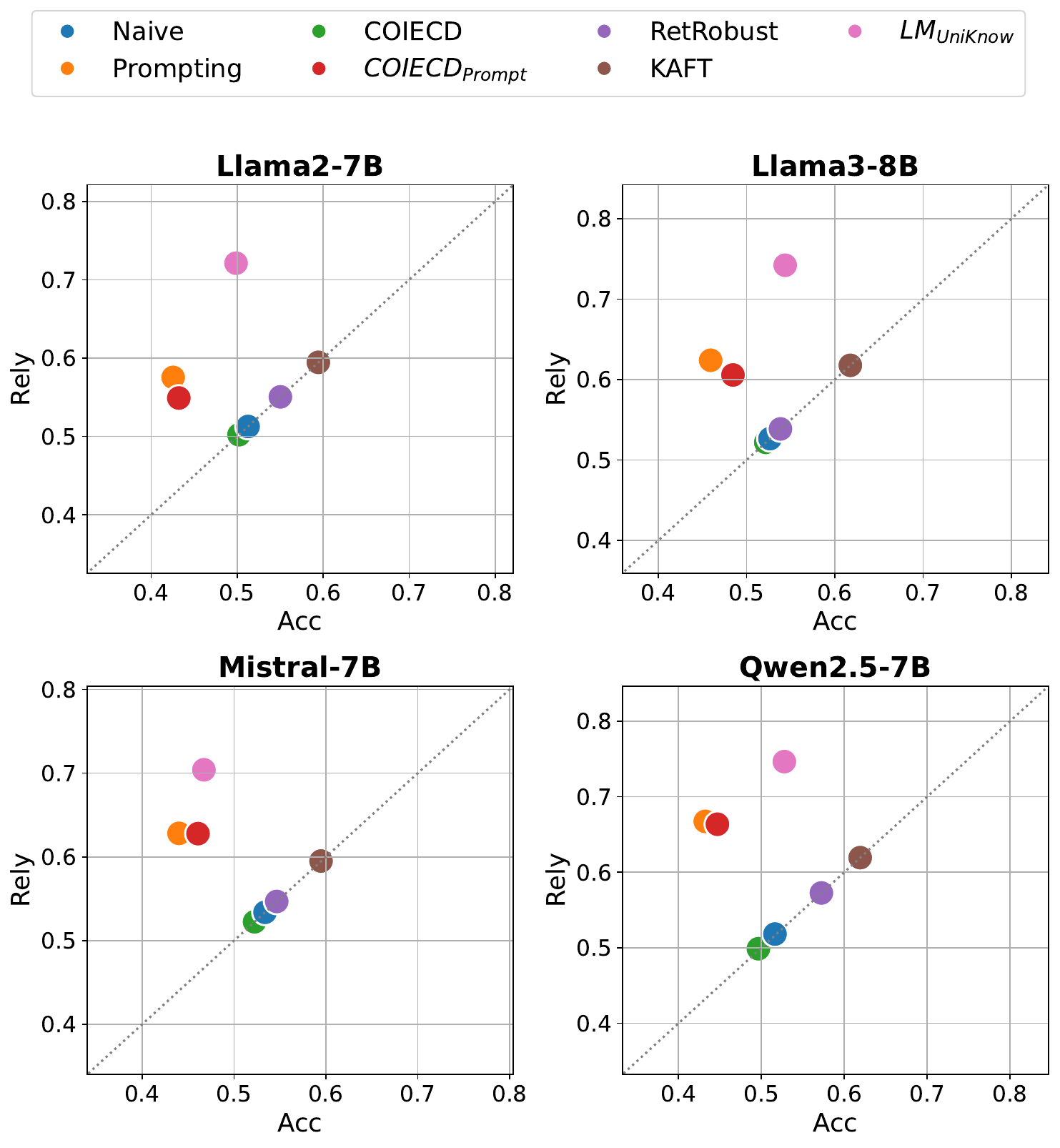}
\caption{\texttt{Acc} and \texttt{Rely} scores across models. Each point represents a method averaged over all datasets. The dotted line indicates equal values of \texttt{Acc} and \texttt{Rely}.}
\label{fig:acc_rely}
\end{figure}

Figure \ref{fig:acc_rely} visualizes the \texttt{Acc} and \texttt{Rely} scores for each method.
Despite including \textit{undefined} samples in the evaluation, the overall trend in \texttt{Rely} scores remains consistent with the scenario-averaged results in \framework\ (\texttt{All} in Figure \ref{fig:framework_figure}).
Note that methods on the dotted line, where \texttt{Acc} equals \texttt{Rely}, limit their performance in terms of answerability.
\ours\ achieves the highest \texttt{Rely}, and its \texttt{Acc} remains comparable to methods which primarily focus on answerability. 
This suggests that, through alignment with \framework, \ours\ effectively minimizes incorrect responses via abstention while maintaining adaptability to various scenarios.

\begin{figure}[t]
\begin{center}
    \includegraphics[width=1\columnwidth]{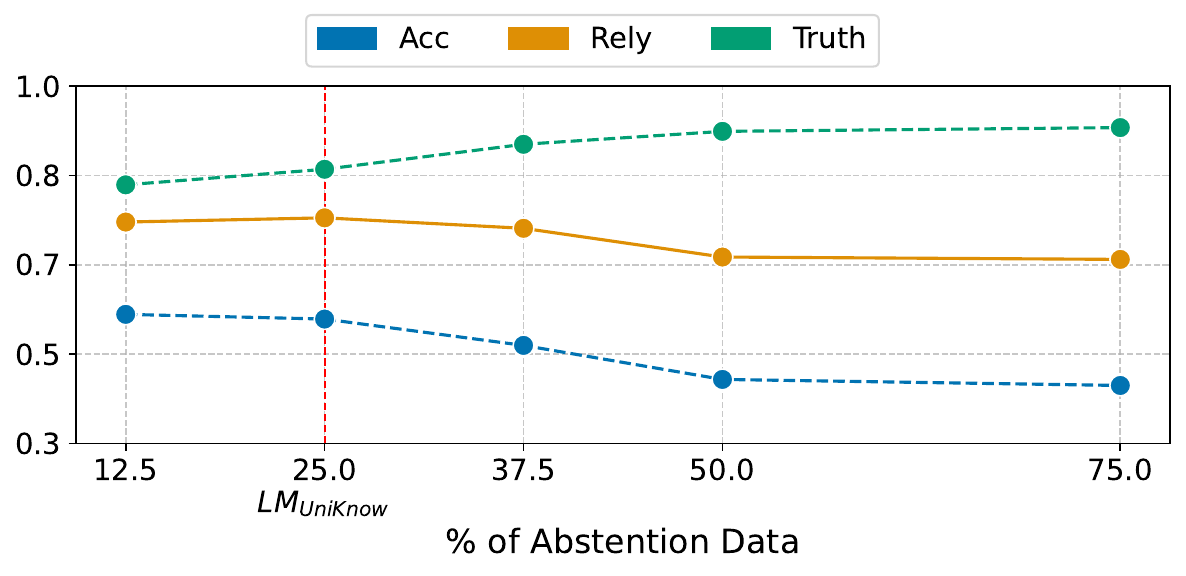}
       \caption{Effect of varying the proportion of abstention data on model performance for {\scshape Llama3-8B}. The red dashed line indicates the proportion used in \ours.} 
      \label{fig:ablation_abstention_avg}
\end{center}
\end{figure}

\begin{table}[t]
\centering
\resizebox{\columnwidth}{!}{
\begin{tabular}{l|ccc|ccc}
\toprule
\multicolumn{1}{r|}{\textbf{Dataset}} & \multicolumn{3}{c|}{\textbf{TriviaQA}} & \multicolumn{3}{c}{\textbf{NQ}} \\
\multicolumn{1}{r|}{\textbf{Metric}} & \textbf{Acc} & \textbf{Truth} & \textbf{Rely} & \textbf{Acc} & \textbf{Truth} & \textbf{Rely} \\
\midrule
\ours           & \textbf{0.6915} & \textbf{0.8762} & \textbf{0.8421} & \textbf{0.5396} & \textbf{0.8161} & \textbf{0.7396} \\
\quad $-$C      & 0.6695 & 0.7040 & 0.7028 & 0.4987 & 0.6430 & 0.6222 \\
\quad $-$IR     & 0.6872 & 0.7352 & 0.7329 & 0.5056 & 0.6410 & 0.6227 \\
\quad $-$C, IR  & 0.6836 & 0.7084 & 0.7078 & 0.4987 & 0.6406 & 0.6205 \\
\bottomrule
\end{tabular}}
\caption{Ablation study on context types in the training data for {\scshape Llama3-8B}, measuring the impact of excluding conflicting contexts ($-$C), incorrectly retrieved contexts ($-$IR), or both ($-$C, IR). \textbf{Bold} indicates the best.}
\label{tab:ablation_context_type}
\end{table}

\subsection{Impact of Abstention Data} \label{sec:abl_abstain}

\ours\ allocates an equal proportion (25\%) to each of the four scenarios within \framework.
To investigate the effect of abstention supervision, we conduct an ablation study using {\scshape Llama3-8B} by varying the proportion of samples from \UI\ scenario.
With a fixed number of training samples, we adjust the proportions of the remaining three scenarios equally.
From Figure \ref{fig:ablation_abstention_avg}, we observe a trade-off between \texttt{Acc} and \texttt{Truth} as the proportion of abstention data increases.
The lower proportions of abstention data lead to higher \texttt{Acc}, while higher proportions improve \texttt{Truth}.
This reflects the inherent trade-off between maximizing correct answer generation (\texttt{Acc}) and minimizing incorrect outputs through abstention (\texttt{Truth}).
Notably, the equal allocation across the four scenarios---25\% abstention data (\ours)---achieves the highest \texttt{Rely} score in both datasets, indicating a balanced performance between answering correctly and abstaining appropriately.

\subsection{Impact of Context Type Diversity}

We conduct an ablation study in which specific types of contexts are selectively removed, while maintaining the total number of training data.
We consider three ablation settings: (1) $-$C, which excludes conflicting contexts and replaces them with original contexts; (2) $-$IR, which removes incorrectly retrieved contexts and retains only randomly sampled irrelevant contexts; and (3) $-$C, IR, which excludes both conflicting and incorrectly retrieved contexts.
These settings allow us to isolate the contribution of each context type to overall reliability.
As shown in Table \ref{tab:ablation_context_type}, excluding conflicting or incorrectly retrieved contexts results in a noticeable drop in \texttt{Truth} and \texttt{Rely}, while having minimal impact on \texttt{Acc}.
These findings underscore the importance of incorporating diverse context types, reflecting those encountered in practical settings, to enhance the reliability of knowledge-handling.

\section{Conclusion}
We present \framework, a unified framework for evaluating LM reliability across PK and EK.
By systematically defining scenarios based on knowledge relevance, \framework\ enables fine-grained analysis of LM behavior.
Our experiments reveal that existing methods often struggle to jointly handle scenarios and exhibit scenario-specific biases.
We show that training with \framework-aligned supervision improves reliability, particularly evident in \UI\ scenario.
\framework\ provides a foundation for building reliable LMs in knowledge utilization.

    \section*{Limitations}

\paragraph{Scope of Knowledge Tasks}
We primarily focus on the QA task, which provides a clear view of knowledge requirements and serves as a representative of knowledge-intensive tasks.
Nevertheless, extending the scope to other tasks--such as reasoning \citep{xiong-etal-2024-large} or claim verification \citep{druid}--is crucial, since the influence of knowledge sources may vary depending on the task.
Additionally, we adopt a simplified RAG setting in which a single context is provided per query, allowing fine-grained control over context relevance and supporting targeted analysis of LM behavior.
However, in real-world applications, LMs often receive multiple retrieved contexts simultaneously.
This introduces new challenges, such as conflicts between external contexts \citep{xu-etal-2024-knowledge-conflicts}.
Incorporating diverse tasks and extending \framework\ to support multi-context would be a valuable step toward modeling more complex and realistic RAG scenarios.

\paragraph{Factuality of External knowledge}
This study assumes that external knowledge is factually accurate, considering scenarios involving changed or newly emerging facts \citep{longpre-etal-2021-entity, xie2023adaptive}.
While this assumption enables controlled analysis, it may be strong in practice, as the quality of external knowledge depends heavily on the underlying database and retrieval system. 
The research area of factuality verification in external contexts using LLMs \citep{yu-etal-2024-truth, fatahi-bayat-etal-2023-fleek} is closely related to this limitation.
Exploring this aspect in conjunction with our framework could further strengthen the setting of the framework.

\paragraph{Limited Strategies for \framework-Aware Training}
Our study focuses on demonstrating the potential to \framework-aware supervised fine-tuning to equip LMs with a comprehensive knowledge utilization capability. 
Still, future work could explore alternative training techniques such as direct preference optimization or reward-based fine-tuning \citep{rafailov2023direct, tian2024finetuning}. 
Broadening the scope of training strategies may offer deeper insights into optimizing LM behavior across scenarios and improve the reliability.
We also leave out trends beyond 14B model scale (e.g. 32B, 70B, 72B), which may further impact behavior in knowledge-intensive tasks.

    \bibliography{anthology,custom}
    
    \clearpage

\appendix
\section*{Appendix}

\section{\framework: Implementation Details}
\label{appendix:implementationDetails}

\subsection{Datasets}

The total number of samples for each dataset is in Table \ref{table:dataset_stat}.
Each sample includes a question, original answer, conflicting answer, and four types of context: original, conflicting, random, and incorrectly retrieved contexts.
We provide a detailed description of the datasets used in our study below.

\paragraph{NaturalQuestions \cite{kwiatkowski-etal-2019-natural}}
Questions consist of real queries issued to the Google search engine.
From a Wikipedia page from the top 5 search results, annotators select a long answer containing enough information to completely infer the answer to the question, and a short answer that comprises the actual answer.
The long answer becomes the context matched with the question, while the short answer is used as the answer.

\paragraph{TriviaQA \cite{joshi2017triviaqalargescaledistantly}}
Question-answer pairs are authored by trivia enthusiasts and independently gathered evidence documents that provide high quality supervision for answering the questions.
The web version of TriviaQA is used, where the contexts are retrieved from the results of a Bing search query.

\paragraph{HotpotQA \cite{yang2018hotpotqadatasetdiverseexplainable}}
Questions are diverse and not constrained to any pre-existing knowledge base. Multi-hop reasoning is required to solve the questions.
Paragraphs that provide supporting facts required for reasoning, are given along with the question.
In the original setting, additional distractor paragraphs are augmented in order to increase the difficulty of inference. However, these distractor paragraphs are not used in this setting.

\paragraph{SQuAD \cite{rajpurkar2016squad100000questionsmachine}}
Paragraphs from Wikipedia are presented to crowdworkers, and they are asked to write questions that entail extractive answers.
The answer to each question is a segment of text from the corresponding reading passage.
To remove the uncertainty that excessively long paragraphs bring, QA pairs that do not align with the first 800 tokens are discarded in this setting.

\paragraph{BioASQ \cite{tsatsaronis2015overview}}
BioASQ is a challenge that assesses the ability of systems to semantically index large numbers of biomedical scientific articles and return concise answers to given natural language questions.
Each question is linked to multiple related science articles. The full abstract of each linked article is used as an individual context. Abstracts that do not exactly contain the answer are discarded.

\paragraph{TextbookQA \cite{Kembhavi_2017_CVPR}}
TextbookQA aims at answering multimodal questions when given a context in formats of text, diagrams and images.
This dataset is collected from lessons from middle school Life Science, Earth Science, and Physical Science textbooks.
Questions that are accompanied with a diagram and "True of False" questions are not used in this setting.

\paragraph{RelationExtraction \cite{levy2017zeroshotrelationextractionreading}}
Given labeled slot-filling examples, relations between entities are transformed into QA pairs using templates. Multiple templates for each type of relation are utilized.
The zero-shot benchmark split of this dataset, which showed that generalization to unseen relations is possible at lower accuracy levels, is used.

\begin{table}[tp]
\centering
\small
\begin{tabular}{@{}r|cc@{}}
\toprule
\textbf{Dataset}            & \textbf{Train} & \textbf{Test} \\ \midrule
\textbf{NQ}                 & 83,787          & 3,994          \\
\textbf{TriviaQA}           & 61,177          & 7,712          \\
\textbf{HotpotQA}           & -              & 4,760          \\
\textbf{SquAD}              & -              & 7,918          \\
\textbf{Bioasq}             & -              & 697          \\
\textbf{TextbookQA}         & -              & 1,056          \\
\textbf{RelationExtraction} & -              & 1,974          \\
\midrule
\textbf{Total}              & 144,964         & 28,111         \\ \bottomrule
\end{tabular}
\caption{Number of samples for each dataset.}
\label{table:dataset_stat}
\end{table}

\subsection{Predefined Abstention Words} \label{app:templates}

\begin{table}[t]
    \centering    
    \small

   \begin{tabularx}{\linewidth}{X}
    \toprule
    
\ttfamily
Answer the following questions: \\
\ttfamily
\textcolor{brown}{<few-shots>} \\
\ttfamily
Question: \textcolor{brown}{<question>} \\
\ttfamily
Answer:
\\
    \bottomrule
\end{tabularx}

    \caption{Template used in closed-book generation.}
    \label{template:cls}
\end{table}

\begin{table}[t]
    \centering    
    \small

   \begin{tabularx}{\linewidth}{X}
    \toprule
    
\ttfamily
Answer the following questions: \\
\ttfamily
\textcolor{brown}{<few-shots>} \\
\ttfamily
Context: \textcolor{brown}{<context>} \\
\ttfamily
Question: \textcolor{brown}{<question>} \\
\ttfamily
Answer:
\\
    \bottomrule
\end{tabularx}

    \caption{Template for the \naive\ open-book generation.}
    \label{template:opn}
\end{table}

\begin{table}[t]
    \centering    
    \small

   \begin{tabularx}{\linewidth}{X}
    \toprule
    
\ttfamily
Answer an entity of the same type as the given keyword. Please note that the keyword is from the given context, and consider the part of speech of the keyword inside the context. You should not give a synonym or alias of the given keyword. The entity and given keyword must have different meanings. Only answer the entity itself without any extra phrases.\\
\ttfamily
\textcolor{brown}{<few-shots>} \\
\ttfamily
Keyword: \textcolor{brown}{<original-answer>} \\
\ttfamily
Context: \textcolor{brown}{<context>} \\
\ttfamily
Answer:
\\
    \bottomrule
\end{tabularx}

    \caption{Template used when instructing the model to generate a conflicting answer, given the original answer and context.}
    \label{template:conflict}
\end{table}

\begin{table}[t]
\centering
\small
\begin{tabularx}{\linewidth}{X}
\toprule
\ttfamily
Answer the following questions. The context may or may not be helpful. If the context is unhelpful and you are not knowledgeable about the question, it is appropriate to say, "\textcolor{brown}{<UNKNOWN>}".\\
\ttfamily
\textcolor{brown}{<few-shots>} \\
\ttfamily
Context: \textcolor{brown}{<context>} \\
\ttfamily
Question: \textcolor{brown}{<question>} \\
\ttfamily
Answer:
\\
    \bottomrule
\end{tabularx}
    \caption{Instruction for LMs to abstain if unknown.}
    \label{template:abs}
\end{table}

\begin{table}[t]
\centering
\small
\begin{tabularx}{\linewidth}{X}
\toprule
\ttfamily
Answer the following questions. If you are not knowledgeable about the question, it is appropriate to say, "\textcolor{brown}{<UNKNOWN>}".\\
\ttfamily
\textcolor{brown}{<few-shots>} \\
\ttfamily
Question: \textcolor{brown}{<question>} \\
\ttfamily
Answer:
\\
    \bottomrule
\end{tabularx}
    \caption{Instruction used in \coiecda\ for LMs to abstain if unknown under closed-book generation.}
    \label{template:abs_cls}
\end{table}

The predefined abstain words \citep{amayuelas2024knowledgeknowledgeexploringknownunknowns} used in evaluations are: [
\texttt{'unanswerable',
    'unknown', 'no known', 'not known', 'do not know'
    'uncertain', 'unclear',
    'no scientific evidence',
    'no definitive answer', 'no right answer', 'no concrete answer',
    'no public information',
    'debate',
    'impossible to know', 'impossible to answer',
    'difficult to predict',
    'not sure',
    'irrelevant', 'not relevant'}]

\subsection{Details on \framework\ Construction}
\label{appendix:datasetConstruction}

To ensure context informativeness and maintain experimental controllability, we have processed the original contexts from the MRQA benchmark by limiting their length and ensuring that the ground-truth answer span is always included.
For each occurrence span of the ground-truth answer in the raw context, we take a 100-word portion surrounding that span and consider it a candidate context.
We then compute the NLI ({\scshape BART-Large}, \citealp{lewis-etal-2020-bart}) score between the question-answer pair and each candidate context, and select the context with the highest NLI score as the original context.

To generate conflicting answers, Template \ref{template:conflict} is employed.
For retrieved-uninformative contexts, a Wikipedia dump from December 2018 is used as a database. Each context is chunked into 100 words. As a retriever model, {\scshape Contriever-msmarco} \citep{izacard2022unsupervised} is utilized.
The number of samples per scenario and model is provided in Table~\ref{tab:scenario_count}.
Template \ref{template:cls} is used to perform closed-book generation for estimating the presence of parametric knowledge.

\section{Knowledge Utilization Methods} 

\subsection{Details on Methods} \label{app:methods}

For \naive\ open-book generation, Template \ref{template:opn} is used.
The instruction template used in the prompting approach is in Template~\ref{template:abs}.

\paragraph{\coiecd}
For \coiecd, which requires two hyperparameters, we adopt the values reported in the original paper ($\alpha=1.0$ and $\lambda=0.25$), as \citet{yuan-etal-2024-discerning} shows that these values generalize well across models and datasets.

\paragraph{\coiecda}
In \coiecda, we use Template~\ref{template:abs} for input with context and Template~\ref{template:abs_cls} for input without context.

\paragraph{\baseline}
Unlike \citealp{li-etal-2023-large}, which treats the parametric answers as gold-standard for irrelevant contexts, we use the original answer to ensure fair evaluation in the \UI\ scenario.

\subsection{Hyperparameters for Training} \label{app:hp_train}
We use the same setting for every training-based approach, including \retrobust, \baseline, and \ours.
We train the model for three epochs using the AdamW \cite{loshchilov2017decoupled} optimizer with a learning rate of 0.0001 and a batch size of 16. For efficient fine-tuning, we employ QLoRA \cite{dettmers2023qlora} with rank r=4 and alpha=16.
Training is conducted on two NVIDIA RTX A6000.

\section{Additional Results}

In this section, we provide exact values of figures and additional results for models not included in Section~\ref{sec:main_results} and Section~\ref{sec:additional}.

\begin{figure}
\centering
\includegraphics[width=\columnwidth]{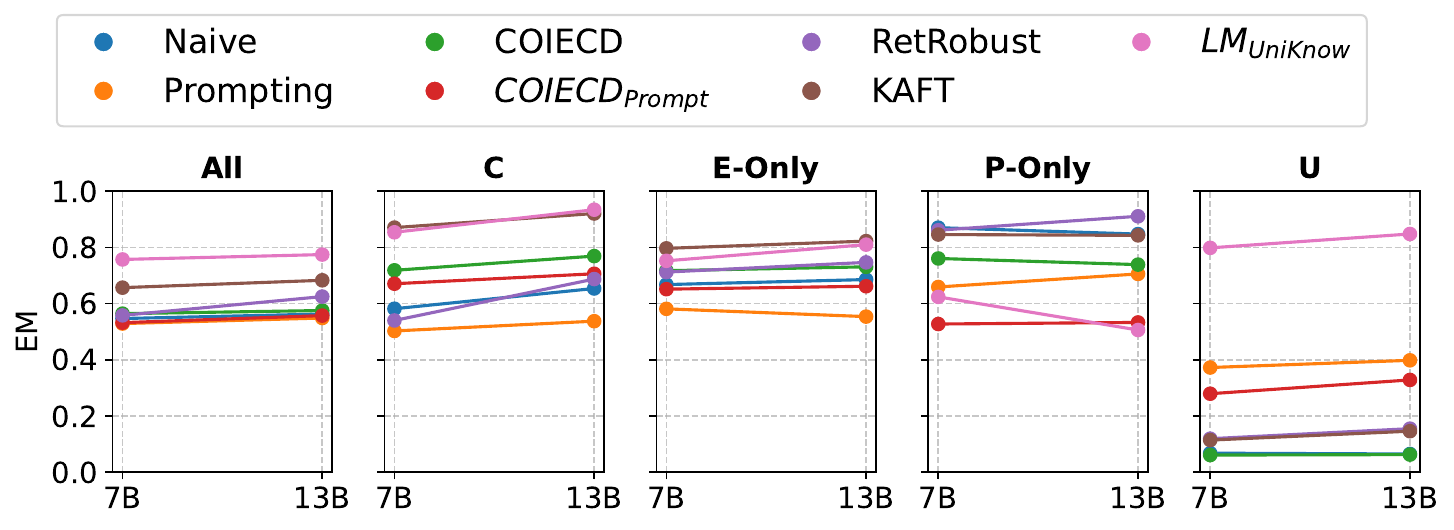}
\caption{EM scores of {\scshape Llama2} models across different sizes, averaged over all datasets within \framework.}
\label{fig:llama2_size}
\end{figure}

\begin{figure}
\centering
\includegraphics[width=\columnwidth]{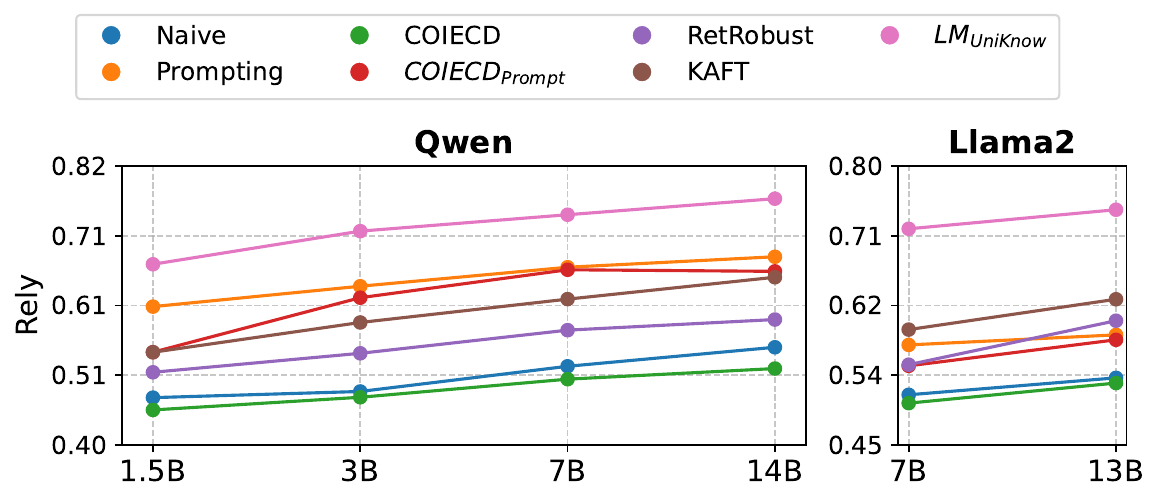}
\caption{\texttt{Rely} scores of {\scshape Qwen} and {\scshape Llama2} across model sizes.}
\label{fig:model_size_rely}
\end{figure}

\paragraph{Main Results}
The EM scores corresponding to Figure \ref{fig:scenario} are provided in Table~\ref{tab:full_scenario}.
Also, Figure \ref{fig:llama2_size} visualizes the EM scores of {\scshape Llama2} 7B and 13B across different knowledge scenarios.
Figure~\ref{fig:model_size_rely} illustrates the impact of model scale with \texttt{Rely} metric for {\scshape Llama2} and {\scshape Qwen2.5}.

\begin{figure}
\centering
\includegraphics[width=\columnwidth]{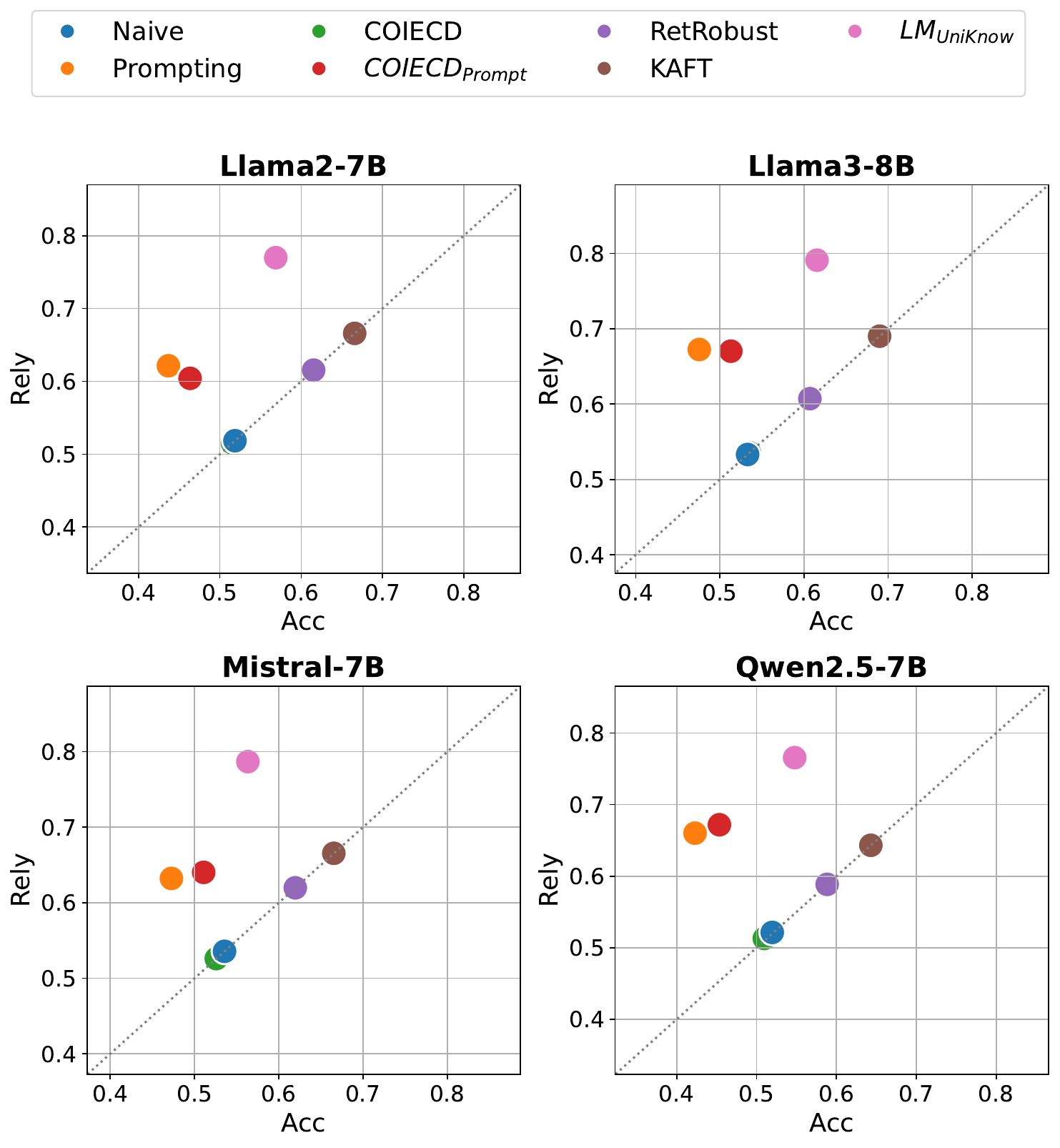}
\caption{\texttt{Acc} and \texttt{Rely} scores averaged over in-domain datasets. Each point represents a method averaged over all datasets. The dotted line indicates equal values of \texttt{Acc} and \texttt{Rely}.}
\label{fig:acc_rely_id}
\end{figure}

\begin{figure}
\centering
\includegraphics[width=\columnwidth]{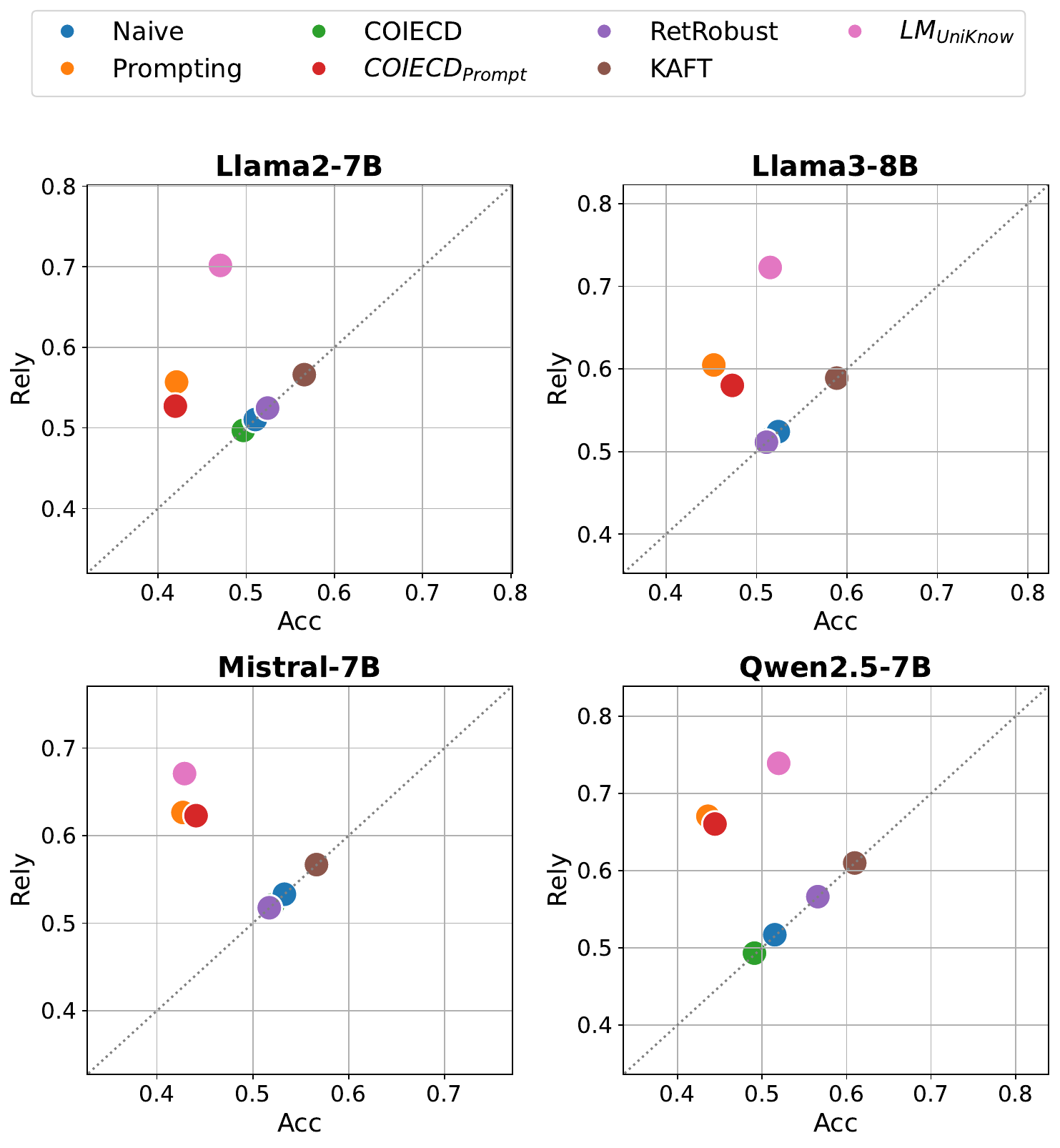}
\caption{\texttt{Acc} and \texttt{Rely} scores averaged over out-of-domain datasets. Each point represents a method averaged over all datasets. The dotted line indicates equal values of \texttt{Acc} and \texttt{Rely}.}
\label{fig:acc_rely_ood}
\end{figure}

The exact values for the \texttt{Acc} and \texttt{Rely} scores presented in Figure~\ref{fig:acc_rely} are listed in Table~\ref{tab:full_Rely} per dataset.
While Figure~\ref{fig:acc_rely} presents overall trends averaged across all datasets, Figure \ref{fig:acc_rely_id} and Figure \ref{fig:acc_rely_ood} break down the results by in-domain and out-of-domain datasets, respectively. 
They further highlight that the overall trend across methods holds consistently and generalizes well to out-of-domain settings.

\begin{figure}
    \centering
    \includegraphics[width=1\columnwidth]{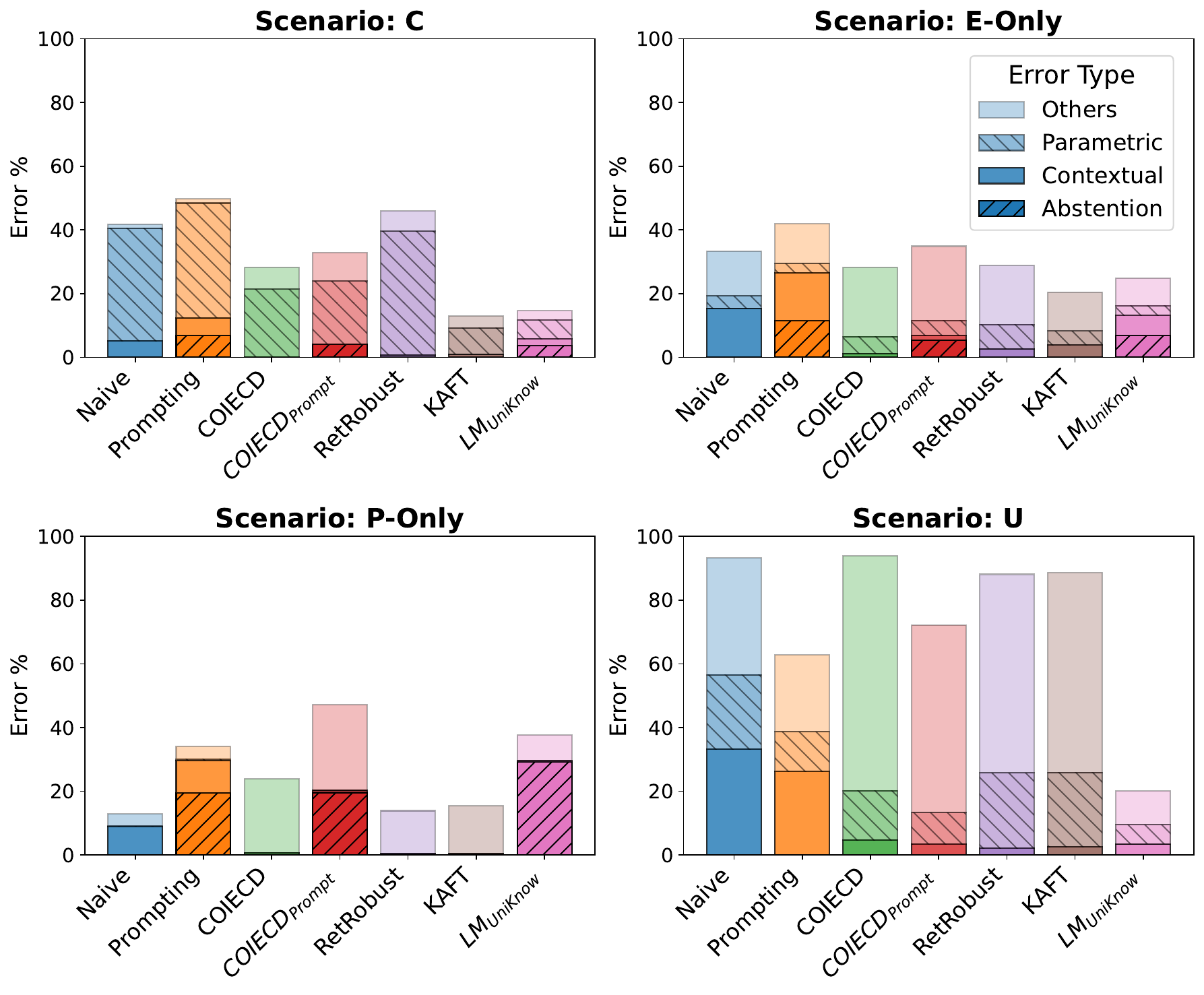} 
    \caption{Stacked error type distributions across methods for each knowledge scenario. Transparency reflects error type. Evaluated using {\scshape Llama2-7B}.}
    \label{fig:error_analysis_llama2}
\end{figure}

\begin{figure}
    \centering
    \includegraphics[width=1\columnwidth]{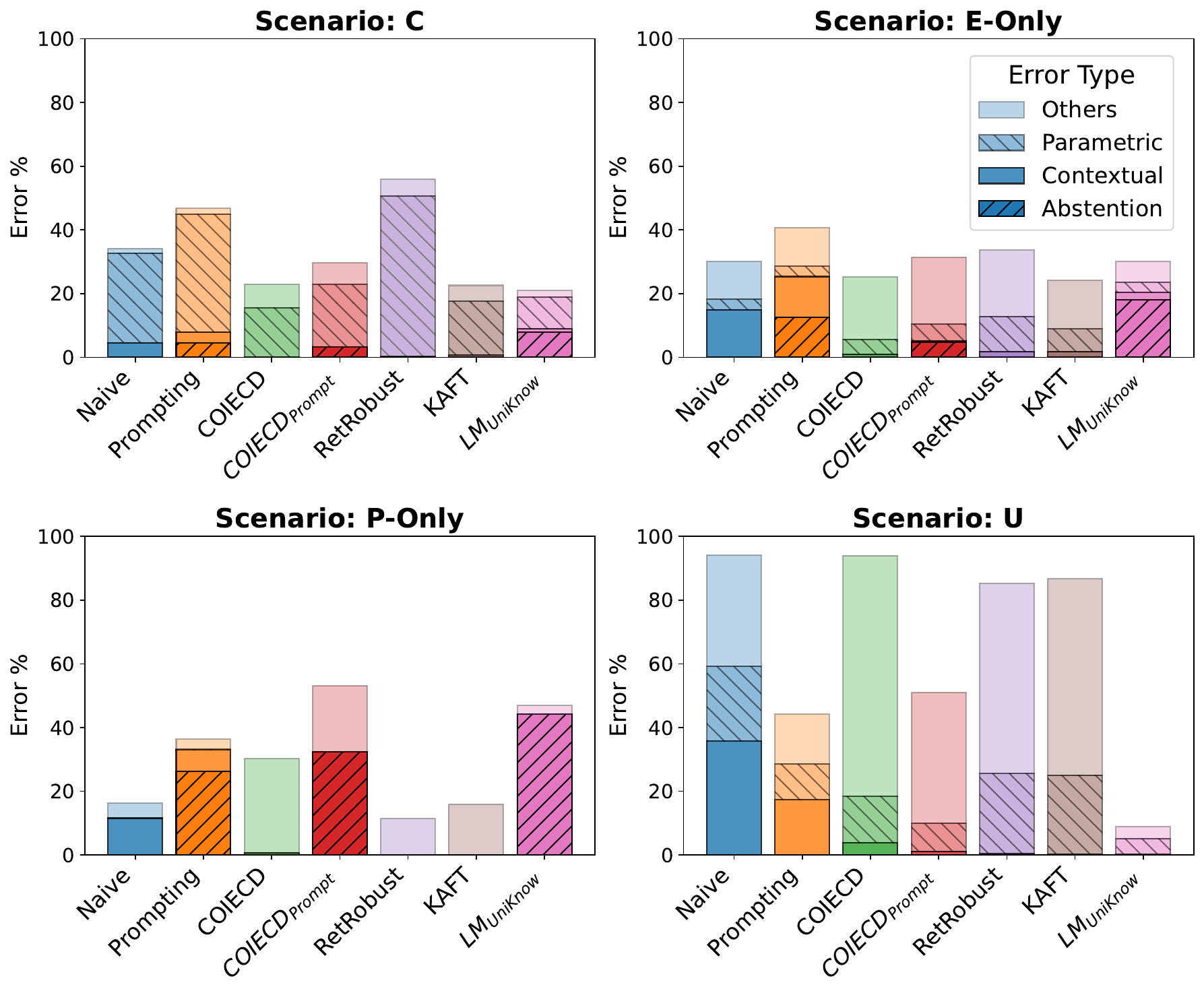} 
    \caption{Stacked error type distributions across methods for each knowledge scenario. Transparency reflects error type. Evaluated using {\scshape Mistral-7B}.}
    \label{fig:error_analysis_mistral}
\end{figure}

\begin{figure}
    \centering
    \includegraphics[width=1\columnwidth]{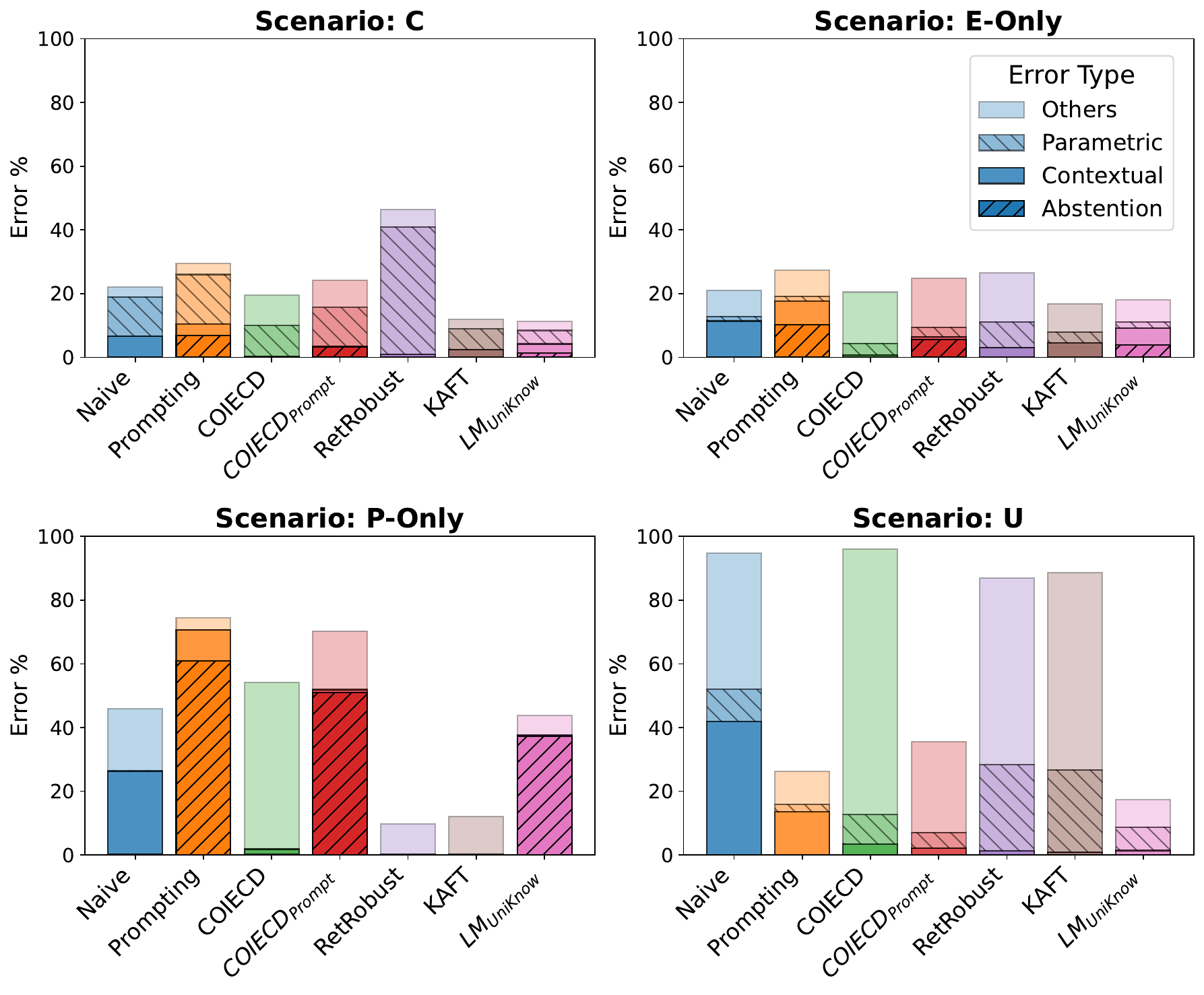} 
    \caption{Stacked error type distributions across methods for each knowledge scenario. Transparency reflects error type. Evaluated using {\scshape Qwen2.5-7B}.}
    \label{fig:error_analysis_qwen}
\end{figure}

\paragraph{Error Analysis}
We present the error type distribution for each knowledge scenario across different models.
Results for {\scshape Llama2-7B}, {\scshape Mistral-7B}, and {\scshape Qwen2.5-7B} are shown in Table \ref{fig:error_analysis_llama2}, Table \ref{fig:error_analysis_mistral}, and Table \ref{fig:error_analysis_qwen}, respectively.

\begin{figure*}
\begin{center}
    \includegraphics[width=1\linewidth]{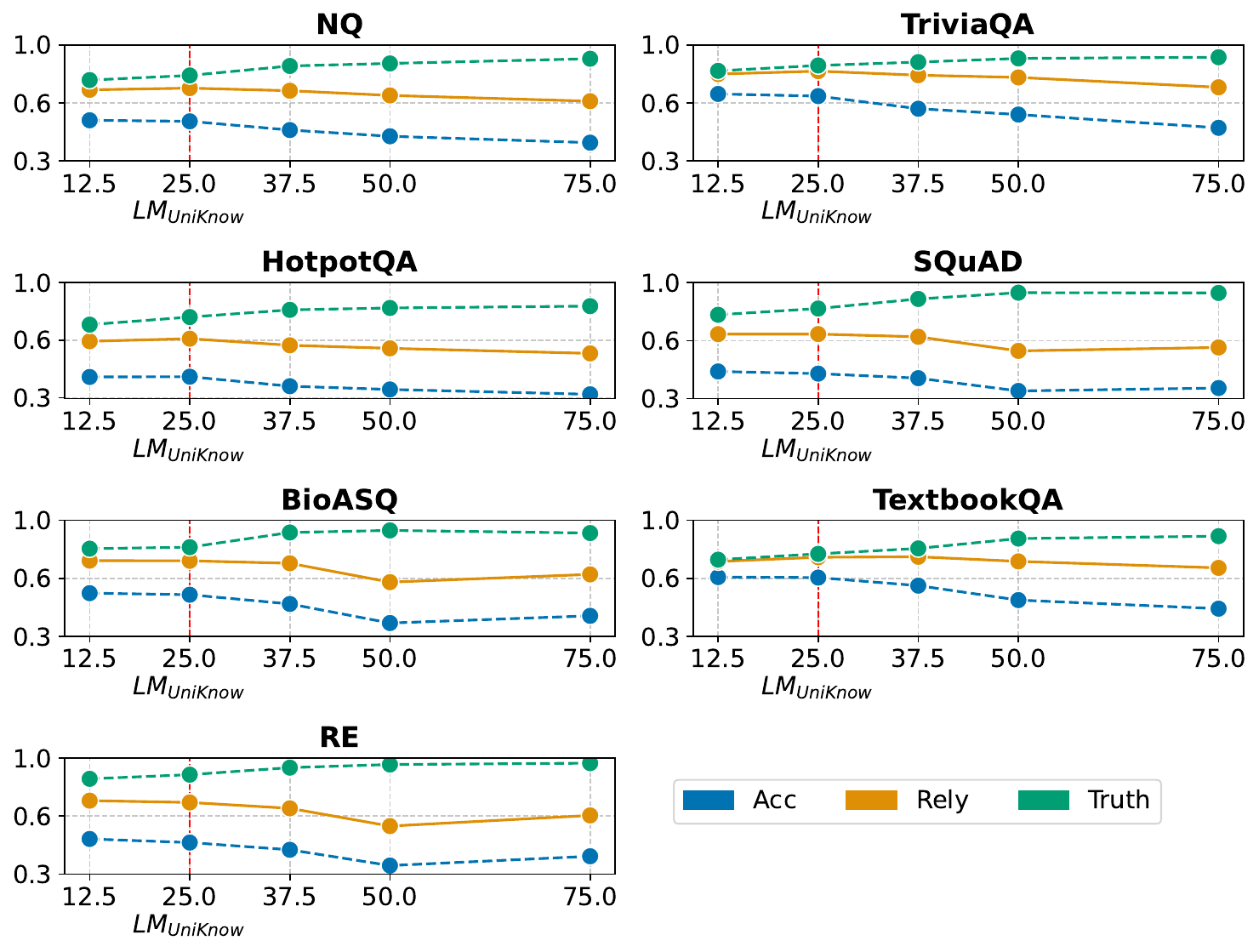}
      \caption{Effect of varying the proportion of abstention data on model performance for {\scshape Llama3-8B} for each dataset. The red dashed line indicates the proportion used in \ours.}
      \label{fig:ablation_abstention_ood}
\end{center}
\end{figure*}

\paragraph{Ablation Study}
Figure~\ref{fig:ablation_abstention_ood} shows the effect of varying the proportion of abstention data on the performance across datasets. 
These results align with the averaged trend discussed in  Section~\ref{sec:abl_abstain}, confirming that the observed pattern holds consistently across datasets.
Table \ref{tab:ablation_context_type_ood} shows the impact of context type diversity on additional datasets beyond those reported in Table~\ref{tab:ablation_context_type}.

\clearpage

\begin{table*}[t]
\centering
\resizebox{\linewidth}{!}{
\begin{tabular}{l|ccc|ccc|ccc|ccc|ccc}
\toprule
\textbf{Dataset} & \multicolumn{3}{c|}{\textbf{HotpotQA}} & \multicolumn{3}{c|}{\textbf{BioASQ}} & \multicolumn{3}{c|}{\textbf{SQuAD}} & \multicolumn{3}{c|}{\textbf{TextbookQA}} & \multicolumn{3}{c}{\textbf{RE}} \\
\textbf{Metric} & Acc & Truth & Rely & Acc & Truth & Rely & Acc & Truth & Rely & Acc & Truth & Rely & Acc & Truth & Rely \\
\midrule
\ours            & \textbf{0.4282} & \textbf{0.7908} & \textbf{0.6593} & 0.5513 & \textbf{0.8379} & \textbf{0.7557} & 0.4513 & \textbf{0.8438} & \textbf{0.6897} & \textbf{0.6546} & \textbf{0.7976} & \textbf{0.7771} & 0.4901 & \textbf{0.8994} & \textbf{0.7319} \\
\quad $-$C       & 0.4506 & 0.4984 & 0.4961 & \textbf{0.5821} & 0.6449 & 0.6410 & 0.4970 & 0.5715 & 0.5659 & 0.6044 & 0.6416 & 0.6402 & 0.5591 & 0.7247 & 0.6973 \\
\quad $-$IR      & 0.4402 & 0.5030 & 0.4990 & 0.5760 & 0.6080 & 0.6069 & \textbf{0.5129} & 0.5565 & 0.5546 & 0.5978 & 0.6089 & 0.6088 & 0.5678 & 0.6331 & 0.6288 \\
\quad $-$C, IR   & 0.4579 & 0.4960 & 0.4946 & 0.5918 & 0.5940 & 0.5940 & 0.5063 & 0.5224 & 0.5221 & 0.6108 & 0.6158 & 0.6157 & \textbf{0.5784} & 0.6312 & 0.6284 \\
\bottomrule
\end{tabular}
}
\caption{
Ablation study on context types in the training data for {\scshape Llama3-8B}, measuring the impact of excluding conflicting contexts ($-$C), incorrectly retrieved contexts ($-$IR), or both ($-$C, IR). \textbf{Bold} indicates the best.
}
\label{tab:ablation_context_type_ood}
\end{table*}

\begin{table*}
\centering
\resizebox{\textwidth}{!}{
\begin{tabular}{llccccccc}
\toprule
Model & Scenario ($\downarrow$) & NQ & TriviaQA & HotpotQA & SQuAD & BioASQ & TextbookQA & RE \\
\midrule
\multirow{4}{*}{{\scshape Llama2-7B}} & \KR &    221 &    2,442 &      160 &     303 &     74 &        175 &    145 \\
        & \KI &    442 &    4,884 &      320 &     606 &    148 &        350 &    290 \\
        & \UR &  5,090 &    3,676 &    6,878 &  11,088 &    626 &        694 &  2,422 \\
        & \UI &  5,090 &    3,676 &    6,878 &  11,088 &    626 &        694 &  2,422 \\
\midrule
\multirow{4}{*}{{\scshape Llama2-13B}} & \KR &    361 &    3,050 &      299 &     431 &     74 &        191 &    207 \\
        & \KI &    722 &    6,100 &      598 &     862 &    148 &        382 &    414 \\
        & \UR &  4,556 &    2,812 &    6,514 &  10,480 &    604 &        632 &  2,306 \\
        & \UI &  4,556 &    2,812 &    6,514 &  10,480 &    604 &        632 &  2,306 \\
\midrule

\multirow{4}{*}{{\scshape Llama3-8B}} & \KR &    273 &    3,231 &      317 &     462 &    101 &        193 &    233 \\
        & \KI &    546 &    6,462 &      634 &     924 &    202 &        386 &    466 \\
        & \UR &  4,766 &    3,076 &    6,444 &  10,360 &    448 &        580 &  2,150 \\
        & \UI &  4,766 &    3,076 &    6,444 &  10,360 &    448 &        580 &  2,150 \\
\midrule
\multirow{4}{*}{{\scshape Mistral-7B}} & \KR &    326 &    3,282 &      302 &     473 &    116 &        220 &    197 \\
        & \KI &    652 &    6,564 &      604 &     946 &    232 &        440 &    394 \\
        & \UR &  4,756 &    3,196 &    6,530 &  10,656 &    494 &        628 &  2,462 \\
        & \UI &  4,756 &    3,196 &    6,530 &  10,656 &    494 &        628 &  2,462 \\
\midrule
\multirow{4}{*}{{\scshape Qwen-1.5B}} & \KR &    119 &    1,011 &       80 &     157 &     59 &        158 &     78 \\
        & \KI &    238 &    2,022 &      160 &     314 &    118 &        316 &    156 \\
        & \UR &  6,202 &    9,246 &    7,774 &  12,292 &    856 &        802 &  2,964 \\
        & \UI &  6,202 &    9,246 &    7,774 &  12,292 &    856 &        802 &  2,964 \\
\midrule
\multirow{4}{*}{{\scshape Qwen-3B}} & \KR &    188 &    1,472 &      167 &     270 &     92 &        184 &    118 \\
        & \KI &    376 &    2,944 &      334 &     540 &    184 &        368 &    236 \\
        & \UR &  5,624 &    7,266 &    7,254 &  11,584 &    580 &        626 &  2,722 \\
        & \UI &  5,624 &    7,266 &    7,254 &  11,584 &    580 &        626 &  2,722 \\
\midrule
\multirow{4}{*}{{\scshape Qwen-7B}} & \KR &    315 &    2,485 &      231 &     401 &    167 &        282 &    187 \\
        & \KI &    630 &    4,970 &      462 &     802 &    334 &        564 &    374 \\
        & \UR &  5,068 &    5,458 &    6,924 &  10,694 &    422 &        502 &  2,460 \\
        & \UI &  5,068 &    5,458 &    6,924 &  10,694 &    422 &        502 &  2,460 \\
\midrule
\multirow{4}{*}{{\scshape Qwen-14B}} & \KR &    334 &    3,284 &      363 &     633 &    202 &        303 &    233 \\
        & \KI &    668 &    6,568 &      726 &   1,266 &    404 &        606 &    466 \\
        & \UR &  4,692 &    3,808 &    6,328 &   9,630 &    316 &        502 &  2,254 \\
        & \UI &  4,692 &    3,808 &    6,328 &   9,630 &    316 &        502 &  2,254 \\
\bottomrule
\end{tabular}}
\caption{Number of samples in each scenario.}
\label{tab:scenario_count}
\end{table*}

\begin{table*}
\centering
\resizebox{\textwidth}{!}{

\begin{tabular}{llcccccccc}
\toprule
Scenario      & Method ($\downarrow$) &           {\scshape Llama2-7B} &          {\scshape Llama2-13B} &           {\scshape Llama3-8B} &          {\scshape Mistral-7B} &           {\scshape Qwen-1.5B} &             {\scshape Qwen-3B} &             {\scshape Qwen-7B} &            {\scshape Qwen-14B} \\
\midrule
\texttt{All} & \Naive &              .5467 &              .5628 &              .5430 &              .5632 &              .5384 &              .5284 &              .5406 &              .5419 \\
               & \absinst &              .5288 &              .5486 &              .5727 &              .5795 &              .5916 &              .5880 &              .6059 &              .6019 \\
               & \coiecd &              .5642 &              .5753 &              .5600 &              .5691 &              .5276 &              .5168 &              .5243 &              .5114 \\
               & \coiecda &              .5321 &              .5572 &              .5881 &              .5870 &              .5516 &              .5819 &              .6130 &              .5976 \\
               & \retrobust &              .5580 &              .6249 &              .5061 &              .5342 &              .5723 &              .5629 &              .5757 &              .5617 \\
               & \baseline &  \underline{.6568} &  \underline{.6829} &  \underline{.6565} &  \underline{.6265} &  \underline{.6344} &  \underline{.6445} &  \underline{.6764} &  \underline{.6831} \\
               & \ours &     \textbf{.7571} &     \textbf{.7745} &     \textbf{.7672} &     \textbf{.7326} &     \textbf{.7120} &     \textbf{.7551} &     \textbf{.7735} &     \textbf{.7918} \\
\midrule
\KR & \Naive &              .5817 &              .6538 &              .5911 &              .6585 &              .7280 &              .7538 &              .7799 &              .7400 \\
               & \absinst &              .5026 &              .5373 &              .5064 &              .5324 &              .7234 &              .7314 &              .7051 &              .6610 \\
               & \coiecd &              .7185 &              .7691 &              .7254 &              .7711 &              .7754 &              .7775 &              .8043 &              .7487 \\
               & \coiecda &              .6707 &              .7061 &              .6979 &              .7033 &              .7591 &              .7515 &              .7587 &              .7112 \\
               & \retrobust &              .5398 &              .6873 &              .3633 &              .4415 &              .6116 &              .5613 &              .5355 &              .5057 \\
               & \baseline &     \textbf{.8706} &  \underline{.9207} &  \underline{.8556} &  \underline{.7741} &  \underline{.8181} &  \underline{.8091} &  \underline{.8818} &  \underline{.9058} \\
               & \ours &  \underline{.8539} &     \textbf{.9343} &     \textbf{.8780} &     \textbf{.7906} &     \textbf{.8290} &     \textbf{.8327} &     \textbf{.8873} &     \textbf{.9287} \\
\midrule
\KI & \Naive &     \textbf{.8703} &  \underline{.8474} &              .8518 &              .8371 &              .6031 &              .5379 &              .5407 &              .5768 \\
               & \absinst &              .6591 &              .7056 &              .7295 &              .6360 &              .4098 &              .2348 &              .2557 &              .3000 \\
               & \coiecd &              .7606 &              .7388 &              .7379 &              .6977 &              .5391 &              .4656 &              .4578 &              .4842 \\
               & \coiecda &              .5272 &              .5329 &              .6051 &              .4685 &              .4152 &              .2888 &              .2975 &              .3447 \\
               & \retrobust &  \underline{.8612} &     \textbf{.9109} &     \textbf{.9081} &     \textbf{.8851} &     \textbf{.8245} &     \textbf{.8522} &     \textbf{.9015} &     \textbf{.8831} \\
               & \baseline &              .8459 &              .8429 &  \underline{.8792} &  \underline{.8407} &  \underline{.8060} &  \underline{.8493} &  \underline{.8795} &  \underline{.8633} \\
               & \ours &              .6239 &              .5059 &              .6713 &              .5299 &              .5549 &              .7233 &              .5620 &              .5668 \\
\midrule
\UR & \Naive &              .6677 &              .6855 &              .6623 &              .6987 &              .7704 &              .7795 &              .7893 &              .7694 \\
               & \absinst &              .5813 &              .5536 &              .5937 &              .5923 &              .7480 &              .7203 &              .7255 &              .7184 \\
               & \coiecd &              .7171 &              .7309 &              .7077 &  \underline{.7478} &              .7594 &              .7841 &              .7952 &              .7580 \\
               & \coiecda &              .6514 &              .6617 &              .6854 &              .6869 &              .7416 &              .7314 &              .7518 &              .7182 \\
               & \retrobust &              .7122 &              .7467 &              .6132 &              .6634 &              .7654 &              .7346 &              .7349 &              .7183 \\
               & \baseline &     \textbf{.7967} &     \textbf{.8225} &  \underline{.7663} &     \textbf{.7583} &     \textbf{.8248} &     \textbf{.8221} &     \textbf{.8314} &  \underline{.8341} \\
               & \ours &  \underline{.7523} &  \underline{.8099} &     \textbf{.7737} &              .6979 &  \underline{.7935} &  \underline{.8171} &  \underline{.8186} &     \textbf{.8408} \\
\midrule
\UI & \Naive &              .0674 &              .0644 &              .0668 &              .0587 &              .0519 &              .0426 &              .0523 &              .0816 \\
               & \absinst &  \underline{.3724} &  \underline{.3980} &  \underline{.4611} &  \underline{.5572} &  \underline{.4852} &     \textbf{.6654} &  \underline{.7371} &  \underline{.7283} \\
               & \coiecd &              .0606 &              .0623 &              .0690 &              .0597 &              .0366 &              .0400 &              .0399 &              .0548 \\
               & \coiecda &              .2790 &              .3282 &              .3641 &              .4891 &              .2904 &              .5560 &              .6442 &              .6164 \\
               & \retrobust &              .1189 &              .1546 &              .1396 &              .1469 &              .0877 &              .1034 &              .1310 &              .1396 \\
               & \baseline &              .1139 &              .1456 &              .1248 &              .1331 &              .0885 &              .0977 &              .1131 &              .1293 \\
               & \ours &     \textbf{.7984} &     \textbf{.8479} &     \textbf{.7460} &     \textbf{.9118} &     \textbf{.6705} &  \underline{.6472} &     \textbf{.8262} &     \textbf{.8308} \\
\bottomrule
\end{tabular}}
\caption{The exact value of EM score for each scenario, across models. \textbf{Bold} indicates the best, and the \underline{underline} indicates the second best.}
\label{tab:full_scenario}
\end{table*}

\begin{table*}[tp]
\centering
\resizebox{\textwidth}{!}{
\begin{tabular}{llcccccccccccccc}
\toprule
Model & Method ($\downarrow$) & \multicolumn{2}{c}{NQ} & \multicolumn{2}{c}{TriviaQA} & \multicolumn{2}{c}{HotpotQA} & \multicolumn{2}{c}{SQuAD} & \multicolumn{2}{c}{BioASQ} & \multicolumn{2}{c}{TextbookQA} & \multicolumn{2}{c}{RE} \\
        &    &                Acc &               Rely &                Acc &               Rely &                Acc &               Rely &                Acc &               Rely &                Acc &               Rely &                Acc &               Rely &                Acc &               Rely \\
\midrule
{\scshape Llama2-7B} & \Naive &              .4177 &              .4177 &              .6194 &              .6194 &              .4342 &              .4342 &  \underline{.4856} &              .4859 &  \underline{.5402} &              .5402 &              .5604 &              .5604 &              .5313 &              .5313 \\
        & \absinst &              .3309 &              .5665 &              .5425 &              .6762 &              .3591 &  \underline{.4675} &              .3748 &              .5134 &              .3849 &              .5776 &              .4799 &              .6318 &              .5067 &  \underline{.5944} \\
        & \coiecd &              .4328 &              .4328 &              .5982 &              .5983 &  \underline{.4355} &              .4356 &              .4818 &              .4822 &              .5147 &              .5147 &              .5284 &              .5284 &              .5234 &              .5236 \\
        & \coiecda &              .3845 &              .5620 &              .5421 &              .6463 &              .3643 &              .4316 &              .3906 &  \underline{.5215} &              .3630 &              .5487 &              .4633 &              .5795 &              .5172 &              .5540 \\
        & \retrobust &  \underline{.5587} &              .5588 &  \underline{.6719} &              .6720 &              .4277 &              .4277 &              .4715 &              .4722 &              .5319 &              .5330 &  \underline{.6241} &              .6241 &  \underline{.5660} &              .5661 \\
        & \baseline &     \textbf{.5990} &  \underline{.5991} &     \textbf{.7327} &  \underline{.7327} &     \textbf{.4485} &              .4486 &     \textbf{.5169} &              .5176 &     \textbf{.5900} &  \underline{.5904} &     \textbf{.6870} &  \underline{.6870} &     \textbf{.5869} &              .5869 \\
        & \ours &              .5167 &     \textbf{.7236} &              .6207 &     \textbf{.8160} &              .3817 &     \textbf{.6219} &              .4401 &     \textbf{.6844} &              .4982 &     \textbf{.7232} &              .5642 &     \textbf{.7628} &              .4695 &     \textbf{.7160} \\
\midrule
{\scshape Llama2-13B} & \Naive &              .4474 &              .4475 &              .6556 &              .6556 &              .4503 &              .4503 &              .5062 &              .5064 &              .5674 &              .5674 &              .5691 &              .5691 &              .5412 &              .5415 \\
        & \absinst &              .3678 &              .5357 &              .5649 &              .7067 &              .3993 &              .4528 &              .4148 &              .6083 &              .2991 &              .5487 &              .5208 &              .6164 &              .4933 &  \underline{.6471} \\
        & \coiecd &              .4594 &              .4594 &              .6361 &              .6362 &              .4509 &              .4510 &              .4959 &              .4961 &              .5739 &              .5739 &              .5533 &              .5535 &              .5222 &              .5223 \\
        & \coiecda &              .4322 &              .5736 &              .6023 &              .6539 &              .3995 &              .4654 &              .4378 &  \underline{.6172} &              .3311 &              .5752 &              .4979 &              .5793 &              .4829 &              .6078 \\
        & \retrobust &  \underline{.6259} &              .6262 &  \underline{.7461} &              .7462 &  \underline{.4752} &              .4753 &  \underline{.5099} &              .5107 &  \underline{.5925} &              .5925 &  \underline{.6889} &              .6892 &  \underline{.5997} &              .6001 \\
        & \baseline &     \textbf{.6445} &  \underline{.6446} &     \textbf{.7890} &  \underline{.7890} &     \textbf{.4934} &  \underline{.4936} &     \textbf{.5448} &              .5457 &     \textbf{.6234} &  \underline{.6248} &     \textbf{.7294} &  \underline{.7294} &     \textbf{.6016} &              .6017 \\
        & \ours &              .5416 &     \textbf{.7510} &              .6348 &     \textbf{.8351} &              .4143 &     \textbf{.6589} &              .4708 &     \textbf{.7145} &              .4598 &     \textbf{.7051} &              .5859 &     \textbf{.7963} &              .5138 &     \textbf{.7548} \\
\midrule
{\scshape Llama3-8B} & \Naive &              .4443 &              .4444 &              .6218 &              .6218 &              .4529 &              .4529 &  \underline{.4943} &              .4944 &              .5656 &              .5656 &              .5627 &              .5627 &              .5447 &              .5447 \\
        & \absinst &              .4200 &  \underline{.6347} &              .5312 &              .7100 &              .3590 &  \underline{.4936} &              .4209 &              .6063 &              .4914 &              .6454 &              .4934 &              .6173 &              .4994 &  \underline{.6613} \\
        & \coiecd &              .4724 &              .4726 &              .5984 &              .5984 &  \underline{.4534} &              .4537 &              .4893 &              .4896 &  \underline{.5857} &              .5864 &              .5301 &              .5301 &              .5230 &              .5230 \\
        & \coiecda &              .4407 &              .6316 &              .5855 &              .7087 &              .3860 &              .4493 &              .4565 &  \underline{.6181} &              .5294 &              .6143 &              .4882 &              .6047 &              .5061 &              .6138 \\
        & \retrobust &  \underline{.5536} &              .5536 &              .6606 &              .6606 &              .4089 &              .4090 &              .4502 &              .4507 &              .5430 &              .5441 &              .6063 &              .6063 &  \underline{.5464} &              .5465 \\
        & \baseline &     \textbf{.6144} &              .6144 &     \textbf{.7657} &  \underline{.7657} &     \textbf{.4764} &              .4764 &     \textbf{.5124} &              .5129 &     \textbf{.6485} &  \underline{.6489} &     \textbf{.7145} &  \underline{.7145} &     \textbf{.5916} &              .5917 \\
        & \ours &              .5396 &     \textbf{.7396} &  \underline{.6915} &     \textbf{.8421} &              .4282 &     \textbf{.6593} &              .4513 &     \textbf{.6897} &              .5513 &     \textbf{.7557} &  \underline{.6546} &     \textbf{.7771} &              .4901 &     \textbf{.7319} \\
\midrule
{\scshape Mistral-7B} & \Naive &              .4444 &              .4444 &              .6270 &              .6270 &  \underline{.4586} &              .4586 &     \textbf{.5109} &              .5111 &  \underline{.5911} &              .5911 &              .5658 &              .5658 &  \underline{.5386} &              .5386 \\
        & \absinst &              .3304 &              .5615 &              .6149 &              .7028 &              .3459 &              .5471 &              .3806 &  \underline{.6145} &              .4634 &  \underline{.6677} &              .4761 &              .6244 &              .4695 &     \textbf{.6786} \\
        & \coiecd &              .4601 &              .4603 &              .5917 &              .5919 &              .4575 &              .4575 &  \underline{.5011} &              .5015 &              .5653 &              .5653 &              .5457 &              .5457 &              .5351 &              .5352 \\
        & \coiecda &              .4039 &              .6053 &              .6179 &              .6750 &              .3525 &  \underline{.5535} &              .4327 &     \textbf{.6389} &              .4516 &              .6269 &              .4967 &              .6267 &              .4705 &  \underline{.6681} \\
        & \retrobust &  \underline{.5873} &              .5875 &  \underline{.6519} &              .6520 &              .4231 &              .4232 &              .4386 &              .4393 &              .5728 &              .5735 &  \underline{.6139} &              .6139 &              .5376 &              .5380 \\
        & \baseline &     \textbf{.6055} &  \underline{.6057} &     \textbf{.7251} &  \underline{.7252} &     \textbf{.4631} &              .4633 &              .4942 &              .4948 &     \textbf{.5983} &              .5994 &     \textbf{.7131} &  \underline{.7131} &     \textbf{.5637} &              .5638 \\
        & \ours &              .5129 &     \textbf{.7466} &              .6138 &     \textbf{.8270} &              .3830 &     \textbf{.6329} &              .3434 &              .5845 &              .4602 &     \textbf{.7071} &              .5533 &     \textbf{.7792} &              .4040 &              .6507 \\
\midrule
{\scshape Qwen-1.5B} & \Naive &              .4300 &              .4306 &              .5005 &              .5014 &              .4056 &              .4057 &  \underline{.4615} &              .4622 &              .4727 &              .4738 &              .5192 &              .5194 &              .5023 &              .5037 \\
        & \absinst &              .4067 &  \underline{.5683} &              .4780 &  \underline{.6333} &              .3723 &  \underline{.5370} &              .4462 &  \underline{.5830} &              .4225 &  \underline{.6465} &              .4427 &              .6174 &              .4547 &  \underline{.6714} \\
        & \coiecd &              .4152 &              .4161 &              .5009 &              .5035 &              .3560 &              .3566 &              .4539 &              .4560 &              .4476 &              .4494 &              .4870 &              .4877 &              .4938 &              .4978 \\
        & \coiecda &              .3769 &              .4919 &              .4845 &              .5681 &              .3383 &              .4280 &              .4297 &              .5218 &              .4362 &              .5668 &              .4657 &              .5653 &              .4743 &              .6341 \\
        & \retrobust &  \underline{.4706} &              .4708 &  \underline{.5502} &              .5503 &  \underline{.4087} &              .4087 &              .4603 &              .4612 &  \underline{.5312} &              .5319 &  \underline{.6113} &              .6115 &  \underline{.5299} &              .5299 \\
        & \baseline &     \textbf{.4904} &              .4905 &     \textbf{.5795} &              .5796 &     \textbf{.4315} &              .4317 &     \textbf{.4904} &              .4913 &     \textbf{.5861} &              .5861 &     \textbf{.6536} &     \textbf{.6539} &     \textbf{.5426} &              .5427 \\
        & \ours &              .4506 &     \textbf{.6414} &              .5342 &     \textbf{.7351} &              .3787 &     \textbf{.6171} &              .4420 &     \textbf{.6547} &              .4878 &     \textbf{.7078} &              .5713 &  \underline{.6323} &              .4783 &     \textbf{.7154} \\
\midrule
{\scshape Qwen-3B} & \Naive &              .4388 &              .4393 &              .5137 &              .5145 &              .4192 &              .4194 &              .4680 &              .4688 &              .4993 &              .4996 &              .5116 &              .5116 &              .5061 &              .5086 \\
        & \absinst &              .3878 &  \underline{.6106} &              .4299 &              .6608 &              .3497 &  \underline{.5903} &              .4279 &  \underline{.6460} &              .4275 &              .6577 &              .4025 &              .6241 &              .4512 &  \underline{.6825} \\
        & \coiecd &              .4263 &              .4278 &              .5130 &              .5174 &              .4066 &              .4075 &              .4617 &              .4636 &              .4803 &              .4831 &              .4858 &              .4889 &              .5048 &              .5125 \\
        & \coiecda &              .3782 &              .5787 &              .4681 &  \underline{.6744} &              .3566 &              .5739 &              .4309 &              .6247 &              .4336 &              .6278 &              .4325 &              .5966 &              .4639 &              .6748 \\
        & \retrobust &              .5134 &              .5135 &              .5804 &              .5805 &              .4251 &              .4251 &              .4688 &              .4698 &              .5864 &              .5868 &              .6477 &              .6477 &  \underline{.5410} &              .5410 \\
        & \baseline &     \textbf{.5397} &              .5399 &     \textbf{.6356} &              .6357 &     \textbf{.4620} &              .4622 &     \textbf{.5106} &              .5114 &     \textbf{.6768} &  \underline{.6786} &     \textbf{.7029} &  \underline{.7029} &     \textbf{.5589} &              .5589 \\
        & \ours &  \underline{.5136} &     \textbf{.6925} &  \underline{.6172} &     \textbf{.7710} &  \underline{.4407} &     \textbf{.6601} &  \underline{.4817} &     \textbf{.6847} &  \underline{.5897} &     \textbf{.7675} &  \underline{.6612} &     \textbf{.7423} &              .5072 &     \textbf{.7349} \\
\midrule
{\scshape Qwen-7B} & \Naive &              .4523 &              .4529 &              .5870 &              .5900 &  \underline{.4413} &              .4450 &              .4905 &              .4929 &              .5735 &              .5742 &              .5573 &              .5578 &              .5132 &              .5147 \\
        & \absinst &              .3788 &              .6230 &              .4672 &              .6973 &              .3759 &  \underline{.6191} &              .4493 &  \underline{.6798} &              .4487 &  \underline{.6860} &              .4422 &              .6659 &              .4639 &  \underline{.7010} \\
        & \coiecd &              .4410 &              .4413 &              .5785 &              .5847 &              .4183 &              .4190 &              .4772 &              .4807 &              .5215 &              .5222 &              .5329 &              .5331 &              .5057 &              .5106 \\
        & \coiecda &              .4096 &  \underline{.6386} &              .4975 &  \underline{.7052} &              .3612 &              .5871 &              .4469 &              .6687 &              .4864 &              .6856 &              .4517 &              .6620 &              .4758 &              .6984 \\
        & \retrobust &  \underline{.5613} &              .5616 &  \underline{.6155} &              .6156 &              .4325 &              .4326 &  \underline{.5139} &              .5147 &  \underline{.6270} &              .6270 &  \underline{.6610} &              .6612 &  \underline{.5959} &              .5959 \\
        & \baseline &     \textbf{.5928} &              .5929 &     \textbf{.6935} &              .6935 &     \textbf{.4772} &              .4773 &     \textbf{.5334} &              .5341 &     \textbf{.6844} &              .6844 &     \textbf{.7502} &  \underline{.7502} &     \textbf{.6033} &              .6033 \\
        & \ours &              .5074 &     \textbf{.7280} &              .5881 &     \textbf{.8032} &              .4083 &     \textbf{.6561} &              .4750 &     \textbf{.7161} &              .5624 &     \textbf{.7768} &              .6409 &     \textbf{.7887} &              .5122 &     \textbf{.7571} \\
\midrule
{\scshape Qwen-14B} & \Naive &              .4743 &              .4746 &              .6522 &              .6523 &              .4614 &              .4616 &              .5051 &              .5063 &  \underline{.6385} &              .6385 &              .5661 &              .5663 &              .5268 &              .5285 \\
        & \absinst &              .4141 &  \underline{.6470} &              .5146 &              .7240 &              .4036 &  \underline{.6293} &              .4510 &  \underline{.6888} &              .4867 &  \underline{.7220} &              .4709 &              .6681 &              .4630 &  \underline{.7035} \\
        & \coiecd &              .4421 &              .4427 &              .6220 &              .6228 &              .4294 &              .4302 &              .4740 &              .4766 &              .5768 &              .5789 &              .5379 &              .5489 &              .5016 &              .5037 \\
        & \coiecda &              .4045 &              .6170 &              .5550 &              .7167 &              .3852 &              .5826 &              .4467 &              .6645 &              .4900 &              .6992 &              .4922 &              .6559 &              .4704 &              .6931 \\
        & \retrobust &  \underline{.6158} &              .6162 &  \underline{.6753} &              .6754 &  \underline{.4741} &              .4741 &  \underline{.5194} &              .5205 &              .5699 &              .5703 &  \underline{.6795} &              .6797 &  \underline{.5845} &              .5845 \\
        & \baseline &     \textbf{.6415} &              .6416 &     \textbf{.7682} &  \underline{.7682} &     \textbf{.5252} &              .5252 &     \textbf{.5637} &              .5646 &     \textbf{.6962} &              .6962 &     \textbf{.7640} &  \underline{.7640} &     \textbf{.6066} &              .6069 \\
        & \ours &              .5453 &     \textbf{.7555} &              .6543 &     \textbf{.8448} &              .4529 &     \textbf{.6912} &              .4926 &     \textbf{.7299} &              .5452 &     \textbf{.7729} &              .6456 &     \textbf{.8335} &              .5342 &     \textbf{.7676} \\
\bottomrule
\end{tabular}}
\caption{Exact values of Acc and Rely for each method and model across datasets. \textbf{Bold} indicates the best, and the \underline{underline} indicates the second best.}
\label{tab:full_Rely}
\end{table*}

\end{document}